\documentclass[10pt,twocolumn,letterpaper]{article}

\usepackage[pagenumbers]{iccv} 

\expandafter\def\expandafter\normalsize\expandafter{%
    \normalsize%
    \setlength\abovedisplayskip{3pt}%
    \setlength\belowdisplayskip{3pt}%
    \setlength\abovedisplayshortskip{-3pt}%
    \setlength\belowdisplayshortskip{3pt}%
}

\usepackage{tabularx}
\definecolor{iccvblue}{rgb}{0.21,0.49,0.74}
\usepackage{multicol}
\usepackage{multirow}
\usepackage[pagebackref,breaklinks,colorlinks,allcolors=iccvblue]{hyperref}
\newcommand\mypar[1]{\par\vspace{0.5mm}\noindent\textbf{#1}\;\;}
\newcommand{\name}{MR. Video\xspace}

\title{MR. Video: ``MapReduce'' is the Principle for Long Video Understanding\vspace{-4mm}}

\author{Ziqi Pang \quad\quad Yu-Xiong Wang \\ University of Illinois Urbana-Champaign}

\begin{document}
\maketitle
\begin{abstract}

We propose \textbf{\name} (pronounced ``mister video''), an agentic long video understanding framework to demonstrate the simple yet effective \textbf{MapReduce} principle for long video understanding: (1) Map, \textbf{independently} and densely perceiving short video clips, and then (2) Reduce, \textbf{jointly} aggregating information from all clips. Compared with sequence-to-sequence vision-language models (VLMs), \name strategically performs detailed short video perception without being constrained by context length. Compared with existing video agents, which typically rely on sequential steps of key segment selection, \name's ``Map'' step enables a simpler and more scalable \textbf{sequence-parallel} perception of short video segments. Meanwhile, its ``Reduce'' step facilitates more \textbf{comprehensive} context aggregation for reasoning, surpassing the explicit key segment retrieval. The ``MapReduce'' principle applies to both VLMs and video agents, and we utilize the convenience of LLM agents to validate its effectiveness. In practice, \name employs two ``MapReduce'' stages: (A) Captioning -- generating short video captions (map), and then standardizing repeated characters and objects into shared names (reduce); (B) Analysis -- for each user question, analyzing relevant information from individual short videos (map), and then integrating them into a final answer (reduce). \name's performance suggests the effectiveness of the MapReduce principle for long video understanding, with \textbf{$>$10\% accuracy improvement} on the challenging LVBench over state-of-the-art VLMs and video agents. The code is at \href{https://github.com/ziqipang/MR-Video}{https://github.com/ziqipang/MR-Video}.

\end{abstract}    
\vspace{-3mm}
\section{Introduction}
\label{sec:intro}
\vspace{-2mm}

\begin{figure}
    \centering
    \includegraphics[width=0.995\linewidth]{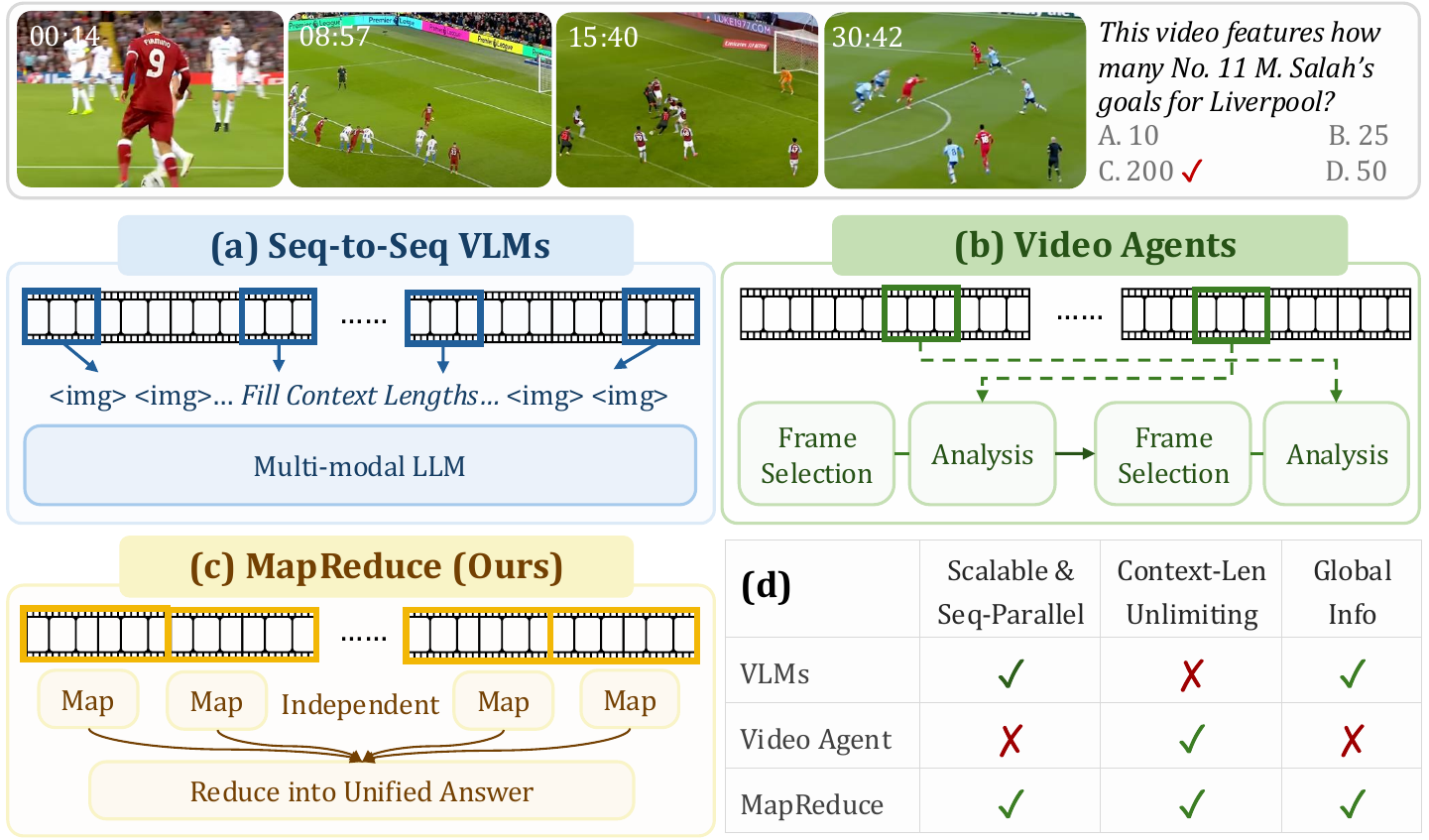}\vspace{-3mm}
    \caption{\textbf{MapReduce Principle.} Long video understanding requires both \emph{global comprehension} and \emph{detailed perception}, as in the motivating example. For such needs, (a) sequence-to-sequence VLMs and (b) video agents are sub-optimal in terms of context lengths, sequential parallelization, and global context information. (c) Instead, we explore and develop a simple ``MapReduce'' principle via \name and (d) overcome such challenges. }
    \vspace{-6mm}
    \label{fig:teaser}
\end{figure}

The ultimate criterion of a long video understanding model is the capability of \emph{digesting global contexts} while \emph{perceiving local details} simultaneously: the model should dedicate itself to all the video contents and not make any assumptions about the video. To clarify this, we show Fig.~\ref{fig:teaser} as a motivating example: in a fast-paced sports highlight \href{https://www.youtube.com/watch?v=i1p5PAgNvR0}{video}, counting the goals of a player requires detailed identification of details for every segment as well as a comprehensive aggregation across the whole video duration. 

Unfortunately, existing sequence-to-sequence vision-language models (VLMs)~\cite{lin2023video, li2024llava, liu2023improvedllava, liu2023llava, liu2024llavanext, zhang2024llavanext-video, li2024llava-interleave} and video agents both have limitations for such scenarios. For instance, VLMs inherently struggle with the contradiction between enormous amounts of long video tokens and limited context lengths, forcing them to conduct frame sampling or token compression (Fig.~\ref{fig:teaser}\textcolor{iccvblue}{(a)}) that might hurt detailed perception, \emph{e.g.}, missing a large portion of ``goals'' in the example. Subsequently, video agents~\cite{wang2024videoagent, fan2024videoagent, yang2024vca} emerge and necessarily bypass VLM's context length via iteratively retrieving the key segments and analyzing their belonging short video clips (Fig.~\ref{fig:teaser}\textcolor{iccvblue}{(b)}). However, they also demonstrate disadvantages compared with VLMs: (1) video agents generally rely on multi-round exploration of video contents, which harms \emph{sequence-parallel} perception of video segments, and one step further, scalability; (2) the explicit key segment retrieval might attend to constrained contexts without \emph{global comprehension} and does not fit into the example requiring a unified information aggregation.

To address these contradictions, we explore a simple yet distinct principle: portraying long video understanding as big data processing, MapReduce~\cite{dean2008mapreduce} suggests an intriguing alternative. (1) ``\emph{Map}'' applies independent and standardized processing of small information units, \emph{i.e.}, short video clips; and (2) ``\emph{Reduce}'' conducts unified reasoning and aggregate unit-level perception into a final answer. If we consider the motivating example of goal counting (Fig.~\ref{fig:teaser}), the MapReduce flow correctly addresses: (1) for VLMs, the limited context lengths only handle a short video clip; (2) for video agents, ``map'' enables independent and sequence-parallel perception of short video clips and ``reduce'' effectively provides global comprehension. Thanks to the convenience of large language models (LLMs), we validate such MapReduce principle into a video agent: ``\textbf{\name}.'' 

We build \name with two MapReduce. (A) \emph{Captioning} stage prepares texts as the efficient medium of video analysis. Its ``map'' step splits the video into atomic scenes and generates dense descriptions. Then, the ``\emph{reduce}'' step enhances consistency by merging the repeated characters or objects from each scene. (B) \emph{Analysis} stage utilizes captions and video frames to answer questions. Its ``map'' generates potential answers or intermediate analysis for atomic short video scenes, and then the ``reduce'' step aggregates the scene-level results into the final answer.  Referring back to the example (Fig.~\ref{fig:teaser}), \name first densely describes the video (Captioning), then counts the scene-level goals and sums them together (Analysis). In fact, we will revisit this example again to examine \name in Sec.~\ref{sec:case_succeed}. 

With the MapReduce principle, our \name significantly pushes the limit of long video understanding: on  LVBench~\cite{wang2024lvbench}, one of the most challenging benchmarks featuring hour-long videos and diverse questions, our \name achieves an accuracy of 60.8\%, \emph{more than 10\% better than previous VLMs and video agents}, with Gemini-Flash perceiving frames and GPT4o understanding texts. 

To summarize, our \name's contributions are:
\begin{enumerate}
    \item We propose the ``MapReduce'' principle for long video understanding, which conceptually mitigates the context length issues of VLMs and sequence-parallel \& global context limitations of video agents.
    \item We design ``\textbf{\name},'' a video agent with simple ``MapReduce'' operations, \emph{i.e.}, sequence-parallel short-video perception and global information aggregation.
    \item We highlight the potential of the ``MapReduce'' principle with \name's significant improvement on benchmarks represented by the challenging LVBench.
\end{enumerate}

We hope the ``MapReduce'' principle, which is verified in LLM agents, is applicable for VLMs can inspire both VLMs and video agents in future works.
\vspace{-2mm}
\section{Related Work}
\label{sec:related}
\vspace{-2mm}
\begin{figure*}
    \centering
    \includegraphics[width=0.99\textwidth]{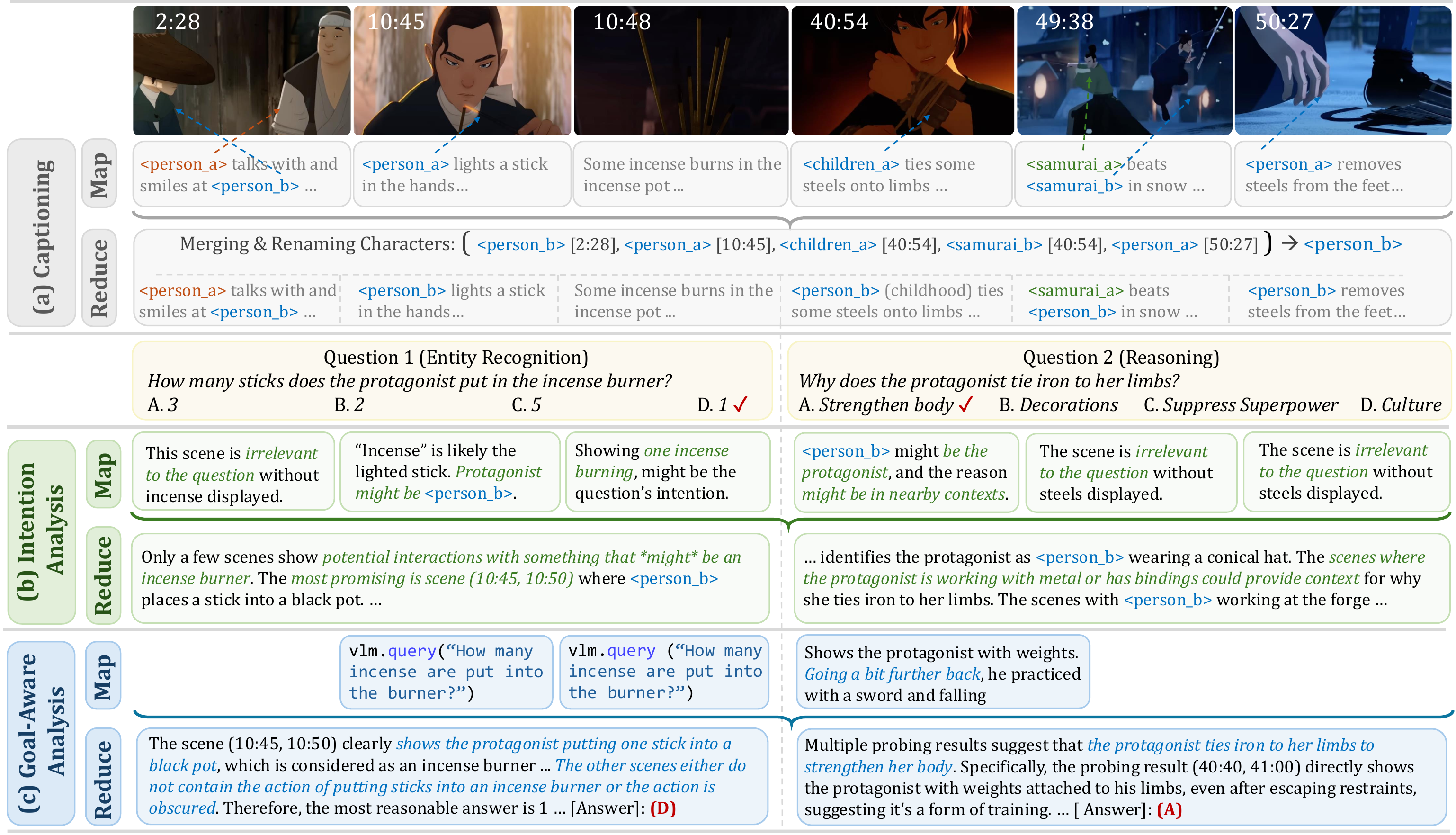}
    \vspace{-3mm}
    \caption{\textbf{Overview.} \name validates the effectiveness of ``MapReduce'' principle with an LLM agent framework. We demonstrate two distinct types of questions for visual details and reasoning. \textbf{(a)} \emph{Captioning} (Sec.~\ref{sec:captioning}) first generates detailed captions of individual scenes (Map) and then enhances consistency by merging repeated characters/objects for the scenes (Reduce). \textbf{(b)} \emph{Question Intention Analysis} (Sec.~\ref{sec:analysis_1}) investigates if a video segment contributes useful information (Map) and then forms a unified analysis at the video level (Reduce). \textbf{(c)} \emph{Goal-Aware Analysis} (Sec.~\ref{sec:analysis_2}) delves deep into detailed perception and reasoning with available contexts, guided by the intention analysis (Map), then unifies them into a final answer (Reduce). For clarity, \name's intermediate texts are simplified.}\vspace{-6mm}
    \label{fig:overview}
\end{figure*}

\mypar{VLMs for Video Understanding.} Existing VLMs~\cite{zhang2024llavanext-video, xue2024longvila, lin2023video, liu2024nvila, li2024videochat, song2024moviechat, shen2024longvu, weng2024longvlm, wang2024longllava, li2024llama, li2023videochat, yang2024qwen2, wang2024qwen2, zohar2024apollo, wang2025internvideo2, li2025temporal, ren2024timechat, shang2024llava, chai2024auroracap, xu2024slowfast, chen2024image, xing2024pyramiddrop, wang2024retake, liu2024oryx} commonly follow LLaVA~\cite{liu2023llava} by projecting image tokens to LLMs. As an image typically takes over 100 tokens in a standard LLaVA model, context lengths become the major challenge for these models in long video understanding: \emph{how to digest the whole video without missing details}? LongVILA~\cite{xue2024longvila}'s solution is increasing the context length, but inherently needs more resources and is still limited by context lengths. Another prevalent solution is decreasing the average tokens per frame via merging or pruning. Such compression can follow certain priors, \emph{e.g.}, similarity of features~\cite{song2024moviechat, shen2024longvu, ren2024timechat, weng2024longvlm, li2024videochat, xu2024slowfast, chen2024image, xing2024pyramiddrop, wang2024retake}, or Q-former-like~\cite{jaegle2021perceiver, jaegl2022eperceiver, li2022blip, li2023blip} learnable module~\cite{li2024llama}. Notably, the recent VideoChat-Flash ~\cite{li2024videochat} can support up to 10k frames with sufficient hardware. However, aggressive compression might lead to unreliable perception of visual details. Such inherent context length limitations of  VLMs necessitate more flexible \emph{agentic paradigms} as below. 

\mypar{Video Understanding Agents.} Video agents provide a meta-level LLM controller on the top of VLMs, which splits a long video into sub-tasks of short video~\cite{fan2024videoagent, zhang2023simple, wang2024videoagent, wang2024videotree, yang2024vca}. Therefore, they are not constrained by context lengths. By imitating how humans watch videos, video agents can be treated as increasing the test-time compute of VLMs via multi-round exploration~\cite{yang2024vca}, key-frame retrieval~\cite{fan2024videoagent}, and tool-use~\cite{wang2024videoagent}. However, video agents still demonstrate disadvantages compared with VLMs, as mentioned in the introduction: (1) the sequential multi-round exploration hinders scalability, and (2) reliance on key-frame retrieval constrains the understanding of sufficient contexts. From such aspects, the ``MapReduce'' principle in our \name bridges these gaps (as in Fig.~\ref{fig:teaser}\textcolor{iccvblue}{(c)}) and marks a distinct and simpler framework for long video agents. 

\mypar{LLM Agents.} Our \name also contributes long video understanding to a broader exploration of addressing complex problems with the advanced reasoning ability of LLM agents, such as software engineering~\cite{jimenez2023swe, yang2025swe} and knowledge retrieval and reasoning~\cite{yang2018hotpotqa, yao2023react, zhou2023language}. The LLM agents can autonomously tackle challenging problems with proper chain-of-thought prompts~\cite{wei2022chain} and decomposed workflows. This work treats LLM agents as a convenient interface to prototype the new ``MapReduce'' principle for long video understanding and indeed shows significant improvement.
\vspace{-2mm}
\section{Method}
\label{sec:method}
\vspace{-2mm}

\subsection{Overview}
\label{sec:overview}
\vspace{-2mm}

As discussed in the introduction (Sec.~\ref{sec:intro}), the ``MapReduce'' principle is compelling for long video understanding because it \emph{conceptually} addresses the challenge of \emph{digesting global contexts} while \emph{perceiving local details}. To verify its effectiveness \emph{in practice}, we validate this principle with our \textbf{\name}: LLM agent provides a convenient way to control the information flow and enables investigation of incorporating ``MapReduce'' into long video understanding.

\name's overview\footnote{The displayed video is the 1st from LVBench (\href{https://www.youtube.com/watch?v=Cm73ma6Ibcs&t=1206s}{video link}). For readers' convenience, we will consistently use it for method demonstrations.} is in Fig.~\ref{fig:overview}. It consists of two major ``MapReduce'' stages. (A) The ``\emph{Captioning}'' stage (Sec.~\ref{sec:captioning}) generates dense captions, which provide a broad comprehension of the video contents and serve as an efficient foundation for answering multiple questions on the same video. (B) The ``\emph{Analysis}'' stage (Sec.~\ref{sec:analysis_1} and Sec.~\ref{sec:analysis_2}) conducts question-specific perception of the video. It first emphasizes understanding the intention of the question (Sec.~\ref{sec:analysis_1}), \emph{i.e.}, ``what the question is \emph{actually} asking,'' then purposefully inspects the visual details or longer temporal spans (Sec.~\ref{sec:analysis_2}). The ``Map'' steps are independent in both stages for different video segments, and the ``Reduce'' stage condenses the segment-level results into unified video-level understanding.

\vspace{-2mm}
\subsection{Captioning}
\label{sec:captioning}
\vspace{-2mm}

Our captioning is shown in Fig.~\ref{fig:overview}\textcolor{iccvblue}{(a)}: (1) The ``\emph{Map}'' step (Sec.~\ref{sec:dense_captioning}) generates dense captions at the scene level, and (2) the ``\emph{Reduce}'' step (Sec.~\ref{sec:reduce_captioning}) provides coherent names for repeated characters and objects for consistency. For 1hr - 2hr videos on LVBench~\cite{wang2024lvbench}, our captioning generates 500 - 2k captions for the whole video, similar to an article.

\vspace{-1mm}
\subsubsection{Map: Dense Scene Captioning}
\label{sec:dense_captioning}
\vspace{-1mm}

We apply sequence-parallel ``Map'' operations for each video segment to generate dense captions and a set of key characters and objects, as in Fig.~\ref{fig:overview}\textcolor{iccvblue}{(a)}.

\mypar{Detailed Description.} \name utilizes VLMs to describe short video clips in detail. We empirically discover that existing VLMs might struggle with processing video clips with multiple actions or significant transitions. Therefore, we first (1) prompt VLMs to check the uniformity of every short video clip\footnote{20 frames in total, 10s segments with frames sampled at 2 fps.} and specify the transitioning frame indexes if the video clip contains multiple scenes; then (2) conduct the captioning task by letting the VLMs describe every \emph{scene} in detail. Such a two-step strategy decreases existing VLMs' difficulties and provides ``scenes'' as atomic units for downstream video perception.

\begin{figure}
    \centering
    \includegraphics[width=0.99\linewidth]{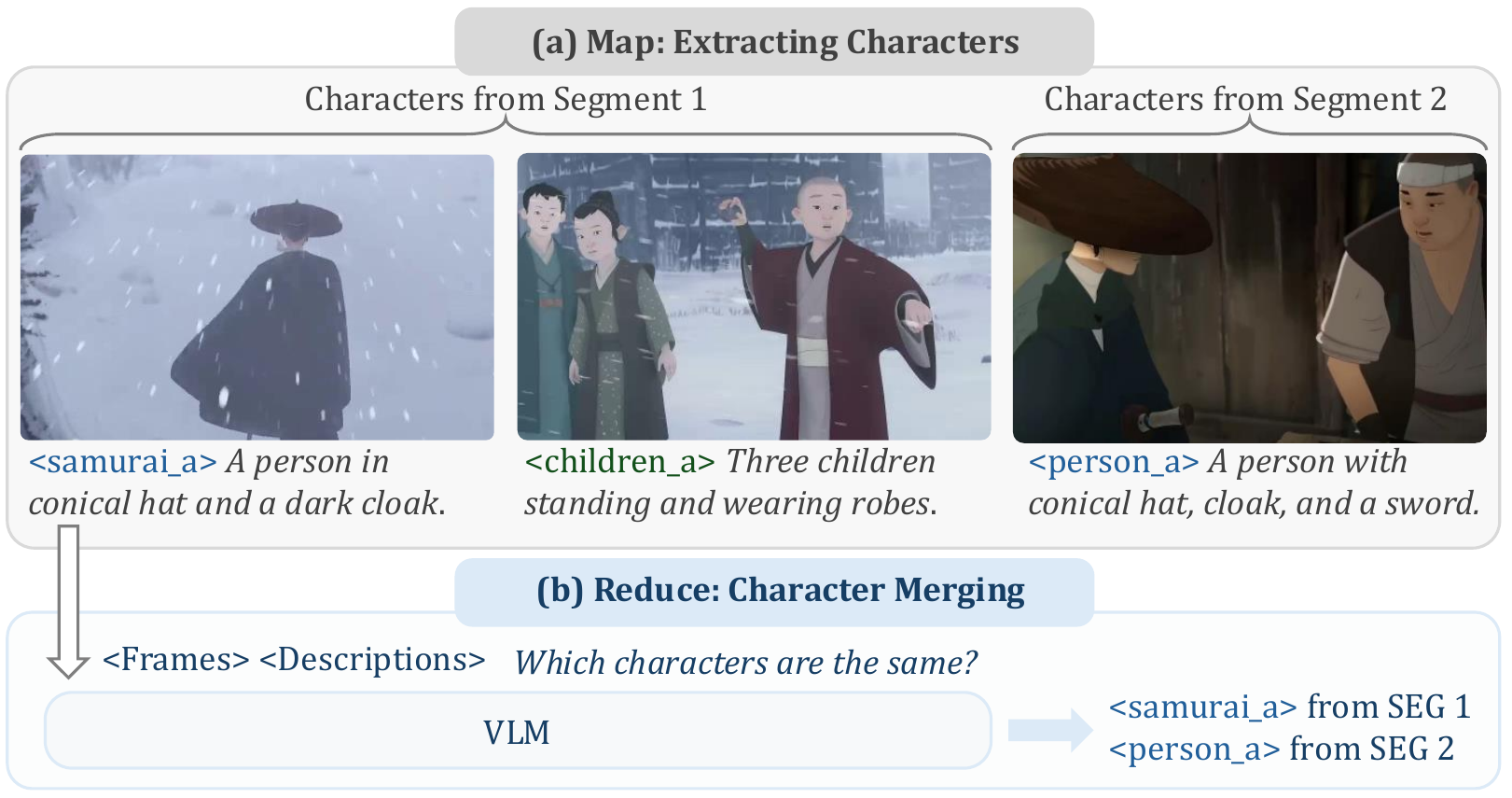}\vspace{-3mm}
    \caption{\textbf{Key Characters/Objects in Captioning.} (a) The ``map'' step extracts the salient characters/objects along with a description, which is useful for frames with multiple characters (the 3rd frame). (b) Then, the ``reduce'' step uses VLM to associate the repeated characters, enhancing the consistency of captions. }
    \vspace{-5mm}
    \label{fig:characters}
\end{figure}

\mypar{Key Characters and Objects.} To enable the object-awareness of captions, we sparsely sample frames from a longer video segment\footnote{30 frames in total, 2min segments with frames sampled at 0.25 fps.} and instruct the VLM to identify the salient characters/objects, describe their identifiable properties, and specify the frame indexes most saliently showing these characters/objects (as Fig.~\ref{fig:characters}\textcolor{iccvblue}{(a)}). These frames, along with their appearance properties, are pre-pended to the contexts of VLMs before generating captions (the 2nd step in the previous paragraph).

\vspace{-1mm}
\subsubsection{Reduce: Consistent Characters and Objects}
\label{sec:reduce_captioning}
\vspace{-1mm}

Identifying consistent characters is an essential requirement for understanding long videos. However, the clip-level captions from ``Map'' steps commonly have inconsistencies due to their sequence-parallel generation, such as ``one name belonging to two characters in different clips'' or ``one character being assigned two names for different video clips,'' as in the ``\emph{Map}'' step of Fig.~\ref{fig:characters}\textcolor{iccvblue}{(a)}. 

To enhance consistency at the video level, our ``\emph{Reduce}'' step merges the repeated characters and resolves contradicting names with \emph{character association} and \emph{caption modification}. (1) \name instructs the VLM to associate the repeated characters/objects by observing the salient frames of extracted characters/objects, as Fig.~\ref{fig:characters}\textcolor{iccvblue}{(b)}. (2) Then \name assigns a new set of names for every character to avoid repeated names and accordingly updates the names in the original clip-level captions to newly created ones. For controllability, we format all the names as ``$<\mathtt{NAME}>$'' in captions. Although using external tracking tools~\cite{fan2024videoagent} might further enhance consistency, we use VLMs because of simplicity and the fact that videos' frequently changing scenes could break the assumption of trackers.

\vspace{-1mm}
\subsection{Analysis I: Question Intention Analysis}
\label{sec:analysis_1}
\vspace{-1mm}

\begin{figure}
    \centering
    \includegraphics[width=0.99\linewidth]{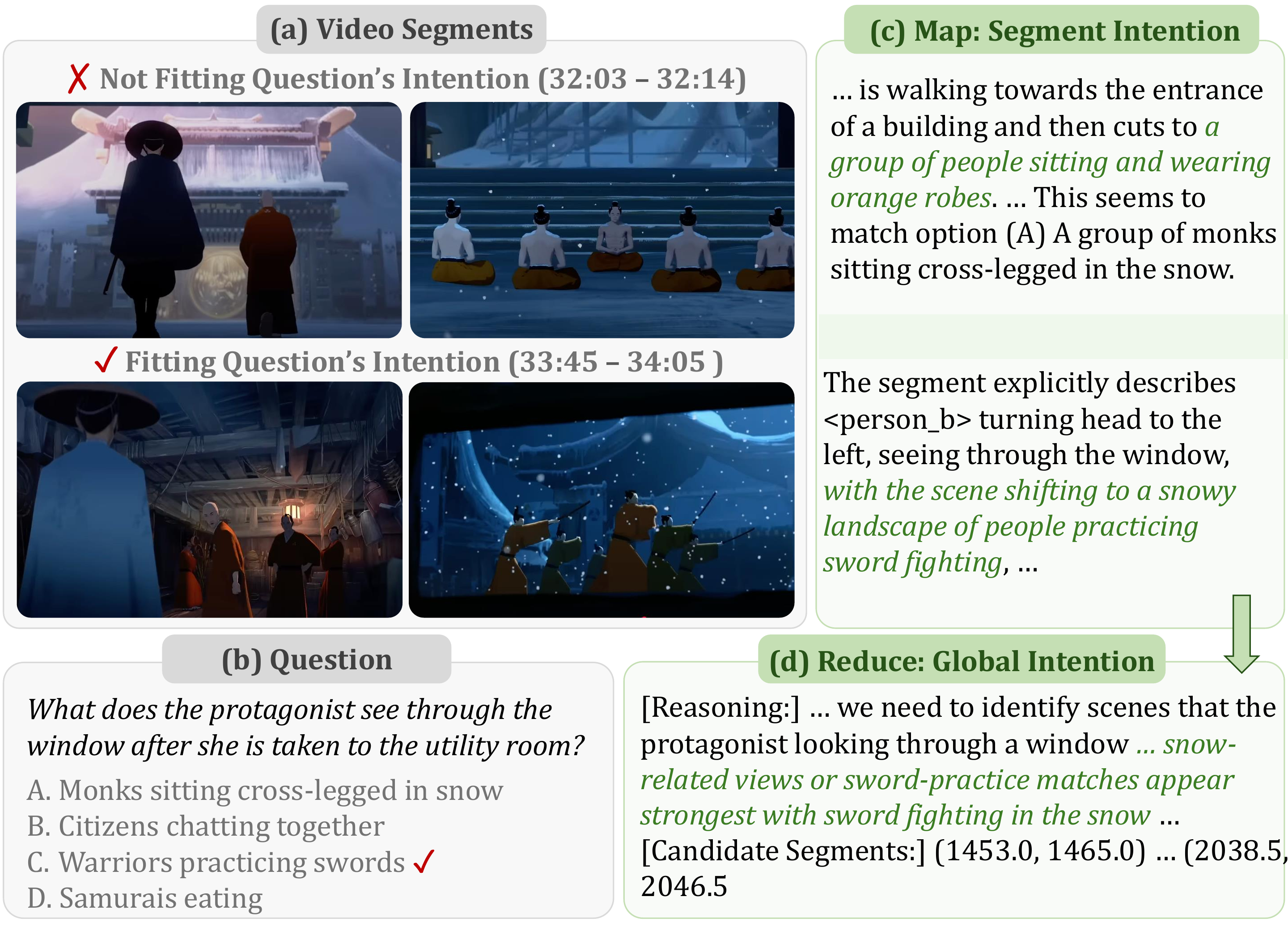}
    \vspace{-3mm}
    \caption{\textbf{Question Intention Analysis.} Long video questions generally require the model to recover certain information, \emph{e.g.}, the meaning of ``protagonist'', what a ``utility room'' looks like, and confounding relevant segments, as in (a). This motivates \name's \emph{explicitly understanding the question's intentions} by reasoning both the video contents and questions.}
    \vspace{-6mm}
    \label{fig:example_intention_analysis}
\end{figure}

\name starts the analysis by understanding the intention of the question. We emphasize the importance of intention analysis because of the inherent \emph{ambiguity of questions} in long-context understanding: the questions might only contain partial information, and the model has to recover crucial clues like ``when,'' ``how long,'' and ``where'' in the video to perceive. For example, Fig.~\ref{fig:example_intention_analysis} demonstrates multiple scenes potentially relevant to the questions, while only one should be correctly selected via reasoning. This stage utilizes the captions from captioning Sec.~\ref{sec:captioning} and optionally includes video frames. Compared with key-frame retrieval in previous video agents~\cite{fan2024videoagent, wang2024videoagent}, \name uniquely combines broader contexts for localizing the relevant frames.

\vspace{-1mm}
\subsubsection{Map: Segment Intention Analysis}
\label{sec:intention_caption_segment}
\vspace{-1mm}
Without losing generality, we divide the video into non-overlapped segments, each empirically containing 32 scenes. For a long video with 1k scenes, this is equivalent to 30 segments. Then, the VLM processes the segments' aggregated captions and middle frames to infer whether any scene is related to the user's question. As each segment is not extremely long with the captions, \name can attend to the details without being limited by the context lengths.

Within each segment, we instruct \name to focus on ``\emph{what is the question asking about}'' and generate a paragraph of analysis as Fig.~\ref{fig:overview}\textcolor{iccvblue}{(b)}. Concretely, its response contains: (1) \emph{Reasoning}: a paragraph analyzing the key subject/criteria mentioned by the questions and how the contents presented in the captions \emph{could} align with the question in any perspective, \emph{e.g.}, Fig.~\ref{fig:example_intention_analysis}\textcolor{iccvblue}{(c)}. (2) \emph{Candidate Scenes}: the LLM then lists the potential scenes that could contribute to answering the question. Please note that this is distinct from directly retrieving key frames since it provides more contexts and allows frames that are helpful in \emph{indirect} ways. (3) \emph{Key Subjects}: The local caption segment becomes insufficient if the question mentions characters or criteria requiring global video information. So \name specifies its unsure criteria and their identifiable properties here for the global ``\emph{Reduce}'' step to analyze.

\vspace{-1mm}
\subsubsection{Reduce: Global Intention Analysis}
\label{sec:intention_global_video}
\vspace{-1mm}

\name's ``reduce'' step aggregates the segment-level analyses at the video level with an LLM processing the texts generated from the ``map'' step. For a 1 hr - 2 hr video, the LLM only needs to handle 10 - 30 paragraphs of the analysis notes, similar to a short article.

The ``reduce'' step generates similar contents as the ``map'' step but covers the contexts of the whole video and localizes the best scenes and subjects for the questions as in Fig.~\ref{fig:overview}\textcolor{iccvblue}{(b)}. Specifically, it responds with: (1) \emph{Reasoning}: a paragraph detailing what the LLM has learned from reading through segment-level analyses and how they fit the questions. (2) \emph{Candidate Scenes}: the ``\emph{reduce}'' step potentially condenses or merges the candidate scenes listed in the ``map'' step according to both video contents and the criteria of questions. (3) \emph{Key Subjects}: by aggregating the global information, LLM can identify the characters/subjects asked by the question and further specify it with the names in captions, \emph{e.g.}, ``$<\mathtt{person}\_{\mathtt{a}}>$.'' An example of the ``reduce'' process is in Fig.~\ref{fig:example_intention_analysis}\textcolor{iccvblue}{(d)}.

\vspace{-1mm}
\subsubsection{Key Segment Retrieval v.s. Intention Analysis}
\vspace{-1mm}

Explicitly reasoning the intention of questions, \emph{i.e.}, completing the contexts, is a significant difference between our MapReduce principle and previous video agents~\cite{fan2024videoagent, wang2024videoagent, yang2024vca}. Our design emphasizes using the models' reasoning abilities to inspect short video clips in detail (map) and then comprehend the video as a whole (reduce). Although the key-segment selection of previous video agents implicitly reflects the ``intention analysis'' objective by choosing a few frames with the most similar features to the question, it is an over-simplified model for long contexts: in the example of Fig.~\ref{fig:example_intention_analysis}, it is challenging to extract features reflecting ``protagonist,'' ``utility room,'' or ``windows'' before understanding the video contexts. Our quantitative analysis on key segment retrieval is in Sec.~\ref{sec:ablation_intention_analysis}.

\vspace{-2mm}
\subsection{Analysis II: Goal-Aware Analysis}
\label{sec:analysis_2}
\vspace{-2mm}

Based on the analyzed question intentions, \name's final MapReduce stage purposefully gathers the information related to the questions and converts them into a final answer, namely, ``\emph{goal-aware analysis},'' as in Fig.~\ref{fig:overview}\textcolor{iccvblue}{(c)}. An essential functionality of this stage is that \name should \emph{explicitly plan the type of information} it needs: attending to captions or sparse frames over longer time horizons provides richer contexts for reasoning, \emph{e.g.}, Q2 in Fig.~\ref{fig:overview}; while focusing on densely sampled frames benefits visual recognition, \emph{e.g.}, Q1 in Fig.~\ref{fig:overview}. With both capabilities, our \name can flexibly handle a wide range of questions.

\begin{figure}
    \centering
    \includegraphics[width=0.99\linewidth]{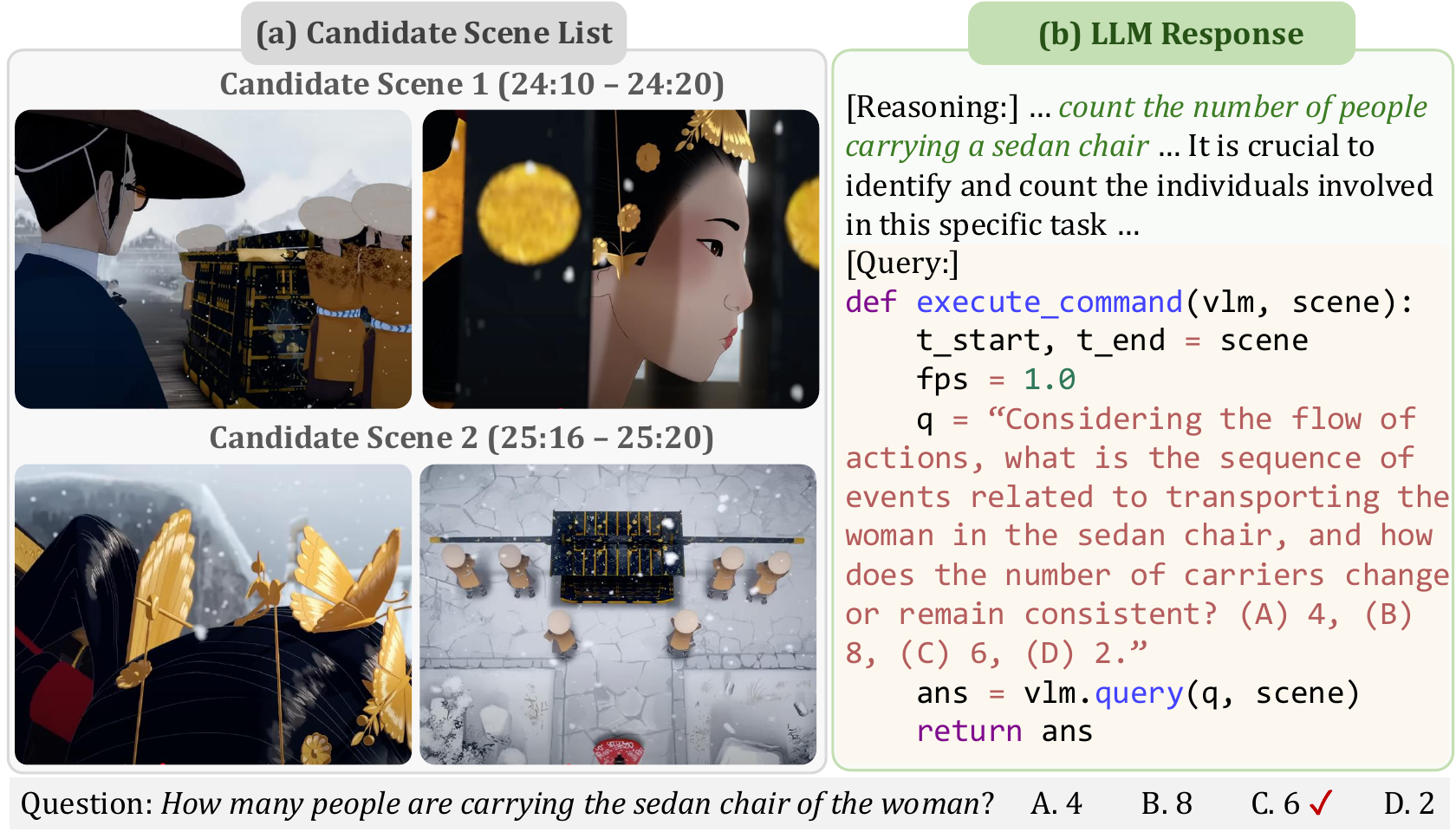}
    \vspace{-3mm}
    \caption{\textbf{Customized Queries for Perception.} With this question requiring detailed visual perception, \name proposes objective-aware queries for the VLMs, confirming the criteria.}
    \vspace{-6mm}
    \label{fig:visprog}
\end{figure}
\vspace{-1mm}
\subsubsection{Map: Goal-Aware Scene-centric Analysis}
\label{sec:scene_centric_analysis}
\vspace{-1mm}

Starting from the candidate scenes generated by question intention analysis (Sec.~\ref{sec:analysis_1}), \name proposes purposeful queries for VLMs to perceive frames \emph{intra-segment} densely sampled frames for visual details or \emph{inter-segment} sparsely sampled frames for global reasoning.

\mypar{Goal Proposal.} When generating the queries for VLMs, we are inspired by the flexibility of ``Visual Programming''~\cite{gupta2023visual} and ViperGPT~\cite{suris2023vipergpt}: let LLM propose its queries for the VLMs and understand the candidate scenes in customized ways. As shown in Fig.~\ref{fig:visprog}, \name proposes the VLM query to cover multiple aspects of the question, gathering comprehensive information.

\mypar{Perceiving Local and Global Information.} We employ the ``reduce'' step as different strategies of sampling frames for VLMs to inspect. (1) \emph{Local}: We densely sample frames within each short segment for objectives requiring detailed visual information, such as the example in Fig.~\ref{fig:visprog}. (2) \emph{Global}: We sparsely sample frames across different segments, \emph{e.g.}, the middle frames of the relevant segments identified by the intention analysis, and let VLM perceive them for information spanning longer temporal ranges. To better leverage the reasoning capabilities of the LLMs, this step can also include the captions of the selected segments to simplify the perception.

\vspace{-1mm}
\subsubsection{Reduce: Answer Generation}\label{sec:answer_gen}
\vspace{-1mm}

The last ``reduce'' step attends to the global context information and generates a final response. With the previous ``map'' steps gradually summarizing the information, this ``reduce'' step is no longer limited by context lengths and can fully unleash reasoning capabilities. As a notable necessity of this ``reduce'' step, it merges the scene analysis results together in a unified way, especially when different scenes provide contradictory perception results or further calculation of scene-level information is required.
\vspace{-2mm}
\section{Experiments}
\label{sec:exp}
\vspace{-1mm}

\subsection{Datasets}
\label{sec:datasets}
\vspace{-1mm}

\mypar{Evaluation Dataset Selection.} To validate the ``MapReduce'' principle within our limited budget, we focus on the challenging long video benchmark: LVBench~\cite{wang2024lvbench}. Compared with others~\cite{mangalam2023egoschema, fu2024video, song2024moviechat, zhou2024mlvu}, LVBench features more extremely long video durations and challenging questions, which are directly reflected by the significantly lowered accuracies of state-of-the-art models. With a limited budget, we expand the breadth of evaluation using the subsets, especially the long video parts of LongVideoBench~\cite{wu2024longvideobench}, Video-MME~\cite{fu2024video}, and EgoShema~\cite{mangalam2023egoschema}.

\mypar{Dataset Settings.} LVbench~\cite{wang2024lvbench} curates 1,549 questions on 103 videos ranging from 30 min to 2 hrs, covering 6 video categories. We utilize the LVBench data as follows. \textbf{(a)} As of March 1st 2025, 4 out of 103 videos are unavailable from YouTube for downloading. So, our comparison in Sec.~\ref{sec:main_comparison} utilizes all the remaining 1,492 questions. \textbf{(b)} For the ablation study (Sec.~\ref{sec:ablation}), we form a subset to save the budget by selecting the first video of each video category in LVBench. This subset has 6 videos and 98 questions in total. For additional evaluation, we use (1) the longest subset of LongVideoBench's validation set, (2) the long video subset of VideoMME without subtitles, and (3) the validation set of EgoSchema. More details are in Sec.~\ref{sec:supp_dataset}.

\vspace{-1mm}
\subsection{Implementation Details}
\label{sec:implementation}
\vspace{-1mm}

\mypar{\name Details.} Our \name demonstrates a simple framework validating the ``MapReduce'' principle, only requiring one LLM for text understanding and one VLM for image understanding. We consistently utilize \textbf{Gemini-2.0-Flash}~\cite{team2023gemini} as our VLM to control our budget and GPT4o as the default LLM. On average, generating the dense captions for an hour-long video requires approximately \$0.8 of Gemini-2.0-Flash, and answering each question from LVbench costs \$0.4 GPT4o on average. We provide further details, especially the prompts, in Sec.~\ref{sec:prompts}.

\mypar{Baseline Evaluation.} For the performance of baselines on evaluated benchmarks, we mainly refer to the numbers on the leaderboards or provided by authors. For Gemini-2.0-Flash, which is our VLM and has not been evaluated on these benchmarks yet, we follow the standard VLM setting by uniformly sampling 256 video frames for long videos (LVBench, LongVideoBench, Video-MME) and 128 frames for the 3min EgoSchema. The prompts are in Sec.~\ref{sec:prompts}.

\mypar{Controlled Context Lengths.} We highlight a vital implementation detail so that our verifying the ``MapReduce'' principle is faithful: \emph{we explicitly control the VLM to perceive less than 32 frames per query}. This ensures \name does not encounter the context length limits.

\begin{table}
\centering
\resizebox{0.99\linewidth}{!}{
\begin{tabular}{l|cccccc|c}
\toprule
Model & ER & EU & KIR & TG & RE & SUM & Overall \\
\midrule
\multicolumn{8}{l}{\emph{Proprietary VLMs}} \\
\midrule
Gemini-1.5-Pro~\cite{team2023gemini} & 32.1 & 30.9 & 39.3 & 31.8 & 27.0 & 32.8 & 33.1  \\
GPT4o~\cite{achiam2023gpt} & 26.5 & 23.7 & 28.3 & 21.4 & 28 & 32.8 & 27.0  \\
Gemini-2.0-Flash~\cite{team2023gemini} & 47.4 & 48.5 & 56.8 & 39.3 & 44.4 & 41.4 & 48.6  \\
\midrule
\multicolumn{8}{l}{\emph{Open-sourced VLMs}} \\
\midrule
InternVL2-40B~\cite{chen2024internvl} & 37.4 & 39.7 & 43.4 & 31.4 & 42.5 & 41.4 & 39.6 \\
TimeMarker~\cite{chen2024timemarker} & 42.8 & 39.1 & 34.9 & 38.7 & 38.2 & 48.8 & 41.3 \\
Qwen2-VL-72B~\cite{wang2024qwen2} & 38.0 & 41.1 & 38.3 & 41.4 & 46.5 & 46.6 & 41.3  \\
VideoLaMA3-2B~\cite{zhang2025videollama} & 41.5 & 39.7 & 44.0 & 32.7 & 45.8 & 25.9 & 41.6  \\
mPLUG-Owl3~\cite{ye2024mplug} & 46.0 & 41.6 & 42.4 & 41.1 & 47.5 & 40.4 & 43.5  \\
InternVL2.5-78B~\cite{chen2024internvl} & 43.8 & 42.0 & 42.1 & 36.8 & 51.0 & 37.9 & 43.6  \\
VideoLLaMA3-7B~\cite{damonlpsg2025videollama3} & 45.8 & 42.4 & 47.8 & 35.9 & 45.8 & 36.2 & 45.3  \\
Qwen2.5-VL-72B~\cite{bai2025qwen2} & - & - & - & - & - & - & 47.7  \\
ReTake~\cite{wang2024retake} & 49.8 & 46.2 & 52.9 & 45.0 & 45.8 & 27.6 & 47.8  \\
VideoChat-Flash~\cite{li2024videochat} & 51.1 & 46.0 & 49.0 & 38.9 & 48.5 & 34.5 & 48.2 \\
GLM-4V-Plus~\cite{glm2024chatglm} & 46.2 & 47.8 & 54.1 & 42.7 & 46.5 & 37.9 & 48.7  \\
\midrule
\multicolumn{8}{l}{\emph{Video Agents}} \\
\midrule
VideoAgent~\cite{wang2024videoagent} & 28.0 & 30.3 & 28.0 & 29.3 & 28.0 & 36.4 & 29.3  \\
VideoTree~\cite{wang2024videotree} & 30.3 & 25.1 & 26.5 & 27.7 & 31.9 & 25.5 & 28.8  \\
VCA~\cite{yang2024vca} & 43.7 & 40.7 & 37.8 & 38.0 & 46.2 & 27.3 & 41.3  \\
\midrule
\name (Ours) & \textbf{59.8} & \textbf{57.4} & \textbf{71.4} & \textbf{58.8} & \textbf{57.7} & \textbf{50.0} & \textbf{60.8} \\
\bottomrule
\end{tabular}
}
\vspace{-3mm}
\caption{\textbf{LVBench Comparison.} Our \name significantly outperforms previous methods by a large $>$10\% margin, suggesting the effectiveness of ``MapReduce'' principle. (Gemini-2.0-Flash is evaluated by ourselves. The VLM accuracies are from the official leaderboard, and the video agent accuracies are from VCA~\cite{yang2024vca}.) }\vspace{-5mm}
\label{tab:lvbench_comparison}
\end{table}

\vspace{-2mm}
\subsection{Main Comparison}
\label{sec:main_comparison}
\vspace{-1mm}
\subsubsection{Nuanced Distinctions of Long Video Benchmarks}
\label{sec:benchmark_nuance}
\vspace{-1mm}

Before analyzing the experimental results, we first clarify an important aspect of long video understanding benchmarks: each benchmark emphasizes different aspects of video understanding in subtle ways. While they all feature a diverse set of questions, they vary in wording and style. As a result, \textbf{none of the existing models show consistent advantage across all the benchmarks}, even GPT4o and Gemini-Pro. 

Comparatively, LVBench and LongVideoBench emphasize the challenges of localizing one or multiple key video clips and \emph{exact} matching of contents, while Video-MME and EgoSchema contain more \emph{interpretative} questions similar to how humans gain an intuitive impression of a video segment. Even LVBench and LongVideoBench are slightly different: LongVideoBench provides more explicit vision-centric cues, and LVBench specifies more from a story or event aspect. More detailed examples and discussion are in Sec.~\ref{sec:dataset_analysis}. Therefore, it is critical to notice the nuances of benchmarks and understand the models comprehensively.

\vspace{-1mm}
\subsubsection{LVBench Comparison}
\vspace{-1mm}

\emph{Given the conceptual advantages of ``MapReduce,'' our experiments verify its practical effectiveness.} We compare \name with VLMs and video agents in Table~\ref{tab:lvbench_comparison}, on the complete set of LVBench~\cite{wang2024lvbench}. As shown in Table~\ref{tab:lvbench_comparison}, our \name shows a significant advantage over all the other methods with an accuracy improvement of over 10\%, indicating a considerable enhancement of long video understanding. As explained in Sec.~\ref{sec:datasets} and Sec.~\ref{sec:benchmark_nuance}, \name features the challenge of connecting the questions with video content. So, \name's effectiveness supports the ``MapReduce'' principle for long video understanding. 

\begin{table}
\centering
\resizebox{0.99\linewidth}{!}{
\begin{tabular}{l|c@{\hspace{2mm}}|c@{\hspace{2mm}}|c@{\hspace{2mm}}|c@{\hspace{2mm}}}
\toprule
\multirow{2}{*}{Model} & LVBench & LongVideoBench & EgoSchema & Video-MME \\
 & Overall & Val (Long) & Val & Long (w/o Sub) \\
\midrule
\multicolumn{5}{l}{\emph{VLMs}} \\
\midrule
Gemini-1.5-Pro~\cite{team2023gemini} & 33.1 & 60.9 & - & \textbf{67.4} \\
GPT4o~\cite{achiam2023gpt} & 27.0 & 58.6 & 70.4 & 65.3 \\
Gemini-2.0-Flash~\cite{team2023gemini} & 48.6 & 45.7 & 71.2 & 63.0 \\
\midrule
\multicolumn{5}{l}{\emph{Video Agents}} \\
\midrule
VideoAgent~\cite{fan2024videoagent} & - & - & 62.8 & - \\
VideoAgent~\cite{wang2024videoagent} & 29.3 & - & 63.2 & - \\
VideoTree~\cite{wang2024videotree} & 28.8 & - & 67.0 & - \\
VCA~\cite{yang2024vca} & 41.3 & - & \textbf{73.6} & - \\
\midrule
\name (Ours) & \textbf{60.8} & \textbf{61.6} & 73.0 & 61.8 \\
\bottomrule
\end{tabular}
}
\vspace{-3mm}
\caption{\textbf{Breadth Comparison on Long Video Benchmarks.} Viewing long video benchmarks comprehensively, as explained in Sec.~\ref{sec:benchmark_nuance}, \name shows better or on-par performance with the baseline models and other video agents. (Except for Gemini-2.0-Flash evaluated by ourselves, LongVideoBench accuracies are from their paper~\cite{wu2024longvideobench}, EgoSchema accuracies are from VCA~\cite{yang2024vca}, and Video-MME accuracies are from the official leaderboard.)}
\vspace{-3mm}
\label{tab:val_comparison}
\end{table}

\vspace{-1.5mm}
\subsubsection{Breadth Comparison}\label{sec:breadth_comparison}
\vspace{-1mm}

As shown in Table~\ref{tab:val_comparison} (LVBench performance is listed for reference), the nuanced differences in benchmarks (Sec.~\ref{sec:benchmark_nuance}) make it challenging for \emph{a single model to dominate all the benchmarks}. However, our \name demonstrates advantages on the LongVideoBench, which requires precise localization of key video segments, similar to LVBench. This further validates our improvement for LVBench and supports the effectiveness of the ``MapReduce'' principle.

On EgoSchema and Video-MME, our \name performs on par with its VLM, Gemini-2.0-Flash, while requiring \emph{a smaller context length} of 32 frames. Although our \name has shown gaps to Gemini-Pro on Video-MME, it is due to a common challenge and future direction of video agents: since the texts are lossy representations of visual contents, video agents naturally struggle with the interpretative questions on a video segment like Video-MME, whereas VLMs are naturally better by observing frames across long ranges without information loss. 

Although training LLMs for text-based video understanding could alleviate this problem compared to zero-shot prompting, it falls outside our objective of validating the ``MapReduce'' principle with simple video agents -- especially given that \name already shows significant improvements on LVBench and LongVideoBench. Therefore, we conclude that \emph{our \name supports the effectiveness and generalization} of the ``MapReduce'' principle.

\vspace{-1mm}
\subsection{Ablation Study and Analysis}
\label{sec:ablation}
\vspace{-1mm}

Since the main objective of \name is to validate the ``MapReduce'' principle, we provide an analysis of its critical steps. By default, we utilize the LVBench subset explained in Sec.~\ref{sec:datasets} for the ablation studies in this section.

\vspace{-1mm}
\subsubsection{Consistent Characters}
\vspace{-1mm}
We first analyze the benefits of the ``\emph{reduce}'' step in captioning (Sec.~\ref{sec:captioning}): providing consistent characters/objects in the captions. Specifically, \name comprehends the captions generated from individual video segments before the ``reduce'' step, and this ``w/o Consistent Characters'' (Fig.~\ref{fig:ablation}\textcolor{iccvblue}{(a)}) performs worse than the \name with consistent characters (59.2\% v.s. 62.2\%). Therefore, aggregating global information, \emph{i.e.}, is essential for long video understanding. From a qualitative sense, consistent characters enable \name to better analyze question intentions. For example, \name can connect different parts of the video and specifies the appearances of ``$<\mathtt{woman}\_{\mathtt{a}}>$'' in Fig.~\ref{fig:visprog} with the help of consistent characters, without only relying on the visual cues of sedan chair. Therefore, the ``reduce'' capabilities enhance the reliability of long video understanding models.

\vspace{-1mm}
\subsubsection{Question Intention Analysis}\label{sec:ablation_intention_analysis}
\vspace{-1mm}

Our question intention analysis is a critical ``MapReduce'' step (Sec.~\ref{sec:analysis_1}) and integrates more context information. We first analyze its effectiveness and comparison with key frame retrieval. The experimental details are in Sec.~\ref{sec:supp_find_relevant_seg_details} (supplementary).

\mypar{Finding Relevant Segments.} With LVBench providing the ground-truth time intervals for each question, we can compare the relevant scenes identified by \name with the annotations of LVBench via recall, where having time overlap is defined as a correct match. \name correctly localizes the relevant scenes for 68.8\% of the questions, where a video typically contains 500-2k scenes for \name to identify from. To illustrate the importance of comprehensive contexts in this process, we compare our question intention analysis with key segment retrieval as below.

\begin{figure}
    \centering
    \includegraphics[width=0.99\linewidth]{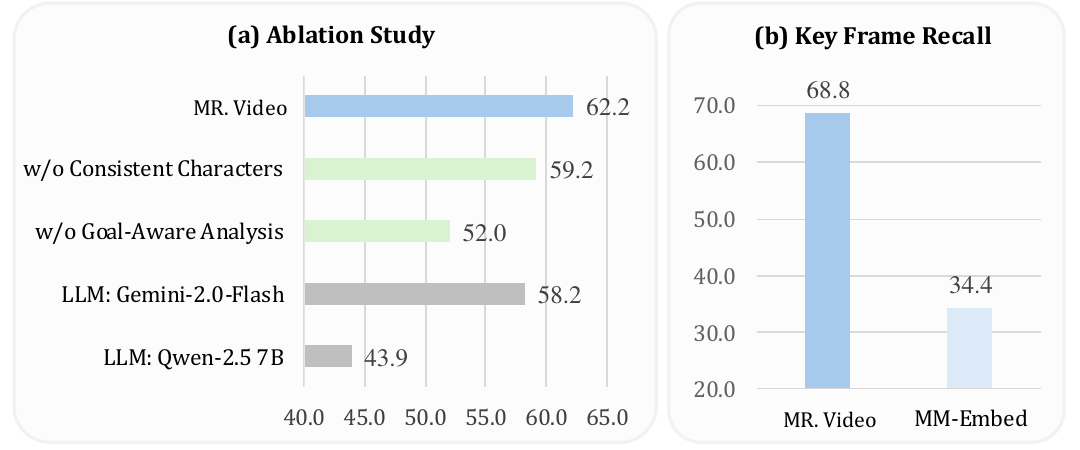}\vspace{-3mm}
    \caption{\textbf{Analytical Experiments of \name}. (a) We investigate the benefits of \name components and the influence of LLMs. (b) The comparison between our question intention analysis and key frame retrieval suggests the necessity of combining more video contexts for understanding.}\vspace{-5mm}
    \label{fig:ablation}
\end{figure}

\mypar{Comparison with Key Segment Retrieval.} We employ MM-Embed~\cite{lin2024mm}, a state-of-the-art multi-modal retrieval model, to encode 256 evenly sampled video frames and the questions for retrieval. Since \name needs to select from 500-2k scenes, the fewer 256 candidates 
decrease the difficulty for the retriever. Following prior works~\cite{wang2024videoagent}, we encode video frames and the questions, then match them according to the inner product of embeddings. Since \name might identify multiple relevant scenes, we let the retriever select the same number of top-k candidate intervals for a fair comparison. Finally, this approach achieves an accuracy of 34.4\%, which is significantly worse than our question intention analysis (Fig.~\ref{fig:ablation}\textcolor{iccvblue}{(b)}). Notably, \name only accesses captions without video frames on LVBench, which proves the necessity of ``reduce,'' \emph{i.e.}, combining global context information, for localizing the key segments.

\vspace{-2mm}
\subsubsection{Goal-aware Analysis}
\vspace{-1mm}

Goal-aware analysis (Sec.~\ref{sec:analysis_2}) provides the video agents with opportunities to delve deep into video content after coarse analysis, a key distinction between video agents and existing VLMs. To validate its necessity, we directly provide the results of global intention analysis to the answer generator (Sec.~\ref{sec:answer_gen}) and evaluate the accuracy of the responses. Such a model achieves an accuracy of 52.0\%, worse than a full \name (Fig.~\ref{fig:ablation}\textcolor{iccvblue}{(a)}). Therefore, it is crucial for perception to inspect the details with goal awareness after understanding the global context.

\vspace{-1mm}
\subsubsection{Different LLMs}
\vspace{-1mm}

LLM plays a crucial role in \name for video reasoning through texts. As shown in Fig.~\ref{fig:ablation}\textcolor{iccvblue}{(a)}, the performance significantly drops if GPT4o is replaced with Gemini-2.0-Flash or Qwen-2.5-7B. Echoing the discussion in Sec.~\ref{sec:breadth_comparison} and Sec.~\ref{sec:case_failure}, it is challenging for an LLM to understand videos via the lossy medium of texts, which is the major bottleneck of video agents. From this perspective, our \name is primarily designed to validate the potential of the ``MapReduce'' principle while improving the language models with training like VLMs would benefit video agents in future works.

\vspace{-2mm}
\subsection{Case Analysis}
\label{sec:case}
\vspace{-2mm}

\subsubsection{Positive Example Analysis}
\label{sec:case_succeed}
\vspace{-1mm}

\begin{figure}
    \centering
    \includegraphics[width=0.99\linewidth]{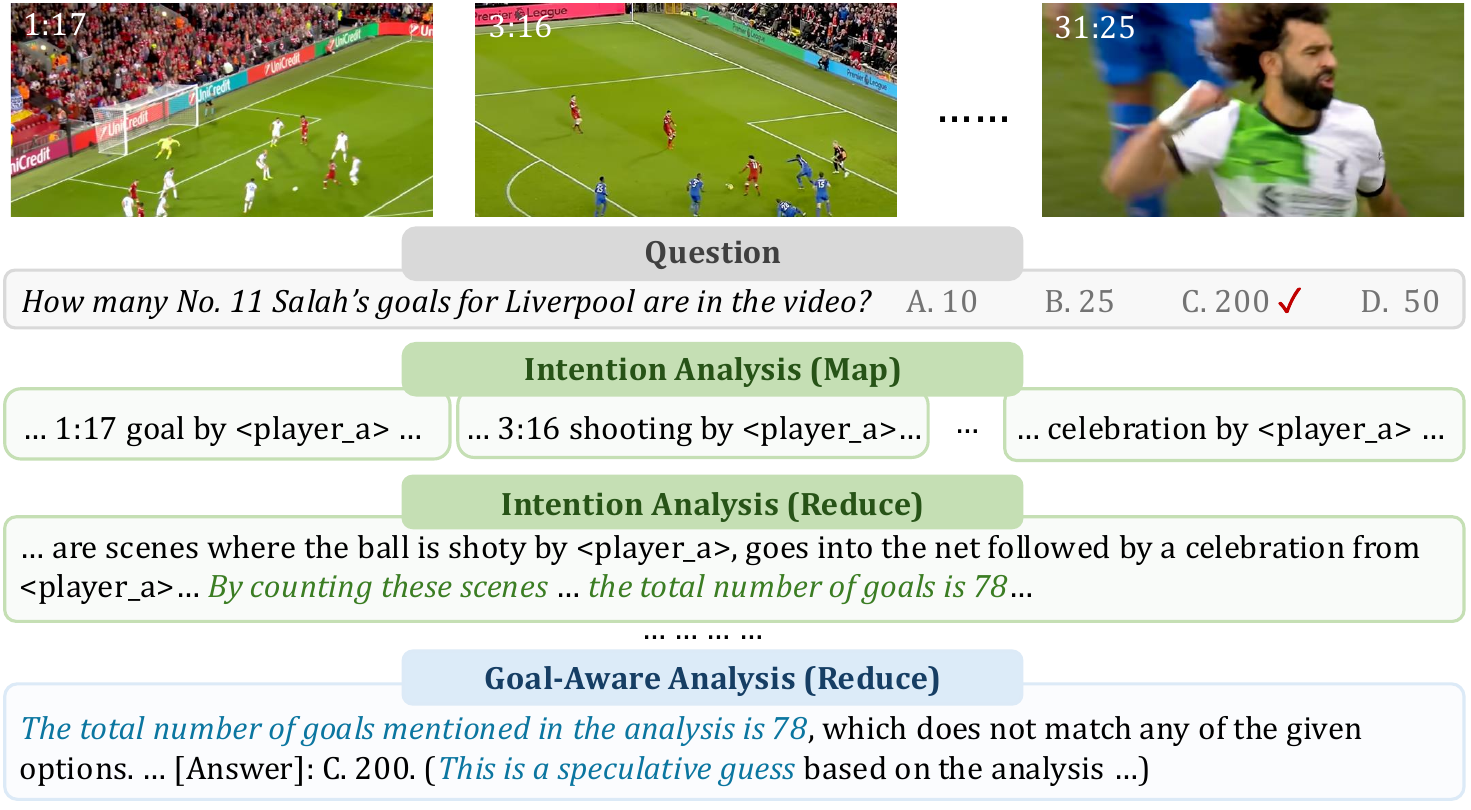}\vspace{-3mm}
    \caption{\textbf{Referring back to the motivating example} (Fig.~\ref{fig:teaser}), \name demonstrates the necessity of ``MapReduce'' principle: checking every scene in detail (map) and aggregating information for the whole video (reduce). Although the model misses some goals due to strict criteria (the full action of shooting, goal, and celebration) and frame sampling, \name shows the desired behavior of \emph{counting exhaustively} and speculates at least 78 goals. }\vspace{-5mm}
    \label{fig:successful_case_counting}
\end{figure}

Finally, we refer back to the motivating example presented in the beginning (Fig.~\ref{fig:teaser}): Is the ``MapReduce'' principle capable of such challenging long video understanding? As shown in Fig.~\ref{fig:successful_case_counting}, \name exhaustively perceives each video clip and searches for the indications of a goal by No. 11; then, the ``reduce'' step counts the number of goals by a rigorous standard. Although the number of goals is smaller than the ground truth due to frame sampling and rigorous standards, \name cannot exclude the wrong answers with the ``MapReduce'' information. Our observations on this challenging case suggest: (1) \name indeed follows the ``MapReduce'' behavior in real scenarios; (2) the ``MapReduce'' principle is instructive for long videos.

\vspace{-2mm}
\subsubsection{Failure Case Analysis}
\label{sec:case_failure}
\vspace{-2mm}

\begin{figure}
    \centering
    \includegraphics[width=0.98\linewidth]{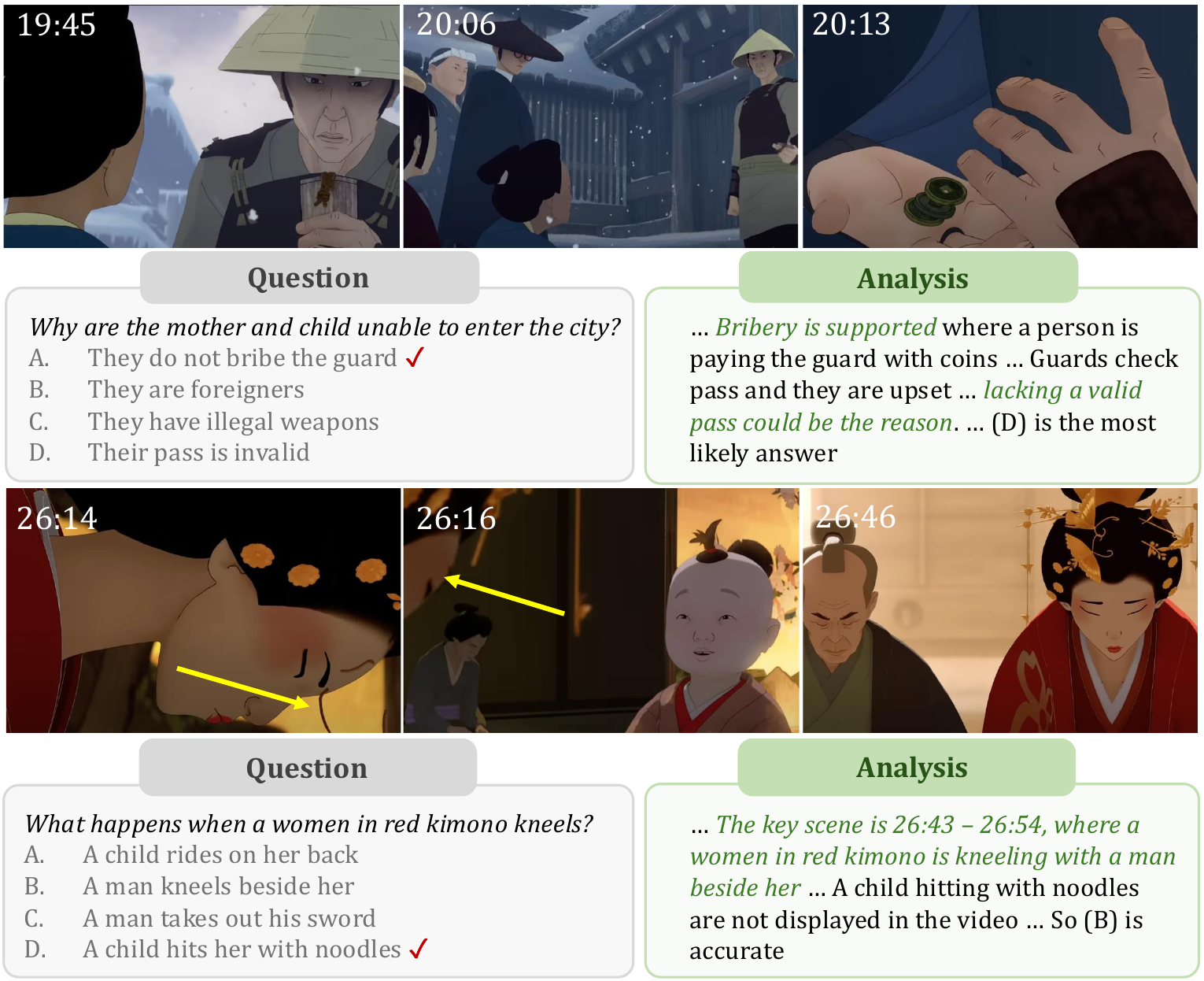}\vspace{-3mm}
    \caption{\textbf{Failure Case Analysis.} \name's failure largely comes from (1) the LLMs \emph{not aligned with human's way of watching videos}, so that it cannot convincingly reason about the scene transitions (example 1); and (2) VLMs failing to capture certain details, where the LLM does not have sufficient information (example 2). (Noodles and the blurry face are pointed with arrows.)}\vspace{-5mm}
    \label{fig:failure_case}
\end{figure}

As demonstrated in the comparisons (Sec.~\ref{sec:main_comparison}), our \name can localize information accurately. However, a major disadvantage and future work of existing video agents is the transitioning between the video and text modalities. (1) The LLMs are trained on language resources, which might not be aligned with \emph{text-based video understanding}. Such misalignment leads to reasoning incapability as in example 1 of Fig.~\ref{fig:failure_case}: the LLM responds with the wrong answer, even though it already notices the right video clip. (2) Moreover, the upper bound of \name is limited by the quality of captions. If the VLM overlooks specific visual details, such as the noodle and woman's face in example 2 of Fig.~\ref{fig:failure_case}, such information loss determines that the LLM cannot get the correct answer. Therefore, both failure analyses reveal the necessity of VLM-like training frameworks for long video understanding instead of relying entirely on the zero-shot abilities of off-the-shelf and VLMs.

\vspace{-3mm}
\section{Conclusions}
\vspace{-2mm}

We introduce a simple and new principle for long video understanding: ``\emph{MapReduce}.'' It conceptually addresses the inherent challenge of long video understanding, \emph{i.e.}, simultaneously \emph{digesting global contexts} while \emph{perceiving local details}. Compared with the frameworks of previous VLMs and video agents, the ``MapReduce'' principle shows the advantage of being free from context length limits, better sequence parallelization, and more comprehensive global contexts. Utilizing the convenience of LLM agents, we validate the effectiveness of ``MapReduce'' with \textbf{\name}: a workflow composed of captioning, question intention analysis, and goal-aware analysis. \name demonstrates an improvement of over 10\% on the challenging LVBench. This firmly pushes the limit of long video understanding and validates the potential of the ``MapReduce'' principle.

\mypar{Limitations and Future Work.} (1) This paper uses LLM agents as a convenient tool to verify the ``MapReduce'' principle, leaving VLMs unexplored. Future work could incorporate this principle into VLMs via techniques like block attention so that highly detailed local perception can happen simultaneously with global context comprehension. (2) As discussed in Sec.~\ref{sec:breadth_comparison} and Sec.~\ref{sec:case_failure}, LLM agents can be enhanced with training or alignment to overcome the information loss issues and enhance the performance for interpretative questions like Video-MME ones.
\clearpage

{
    \small
    \bibliographystyle{ieeenat_fullname}
    \bibliography{main}
}

\clearpage
\maketitlesupplementary

\renewcommand\thesection{\Alph{section}}
\renewcommand\thetable{\Alph{table}}
\renewcommand\thefigure{\Alph{figure}}
\renewcommand\theequation{\Alph{equation}}
\setcounter{section}{0}
\setcounter{table}{0}
\setcounter{figure}{0}
\setcounter{equation}{0}

\section{Delving into Long Video Benchmarks}
\label{sec:dataset_analysis}

In Sec.~\ref{sec:benchmark_nuance}, we discussed the critical fact that the long video datasets have nuanced differences in their preferred long video understanding capabilities. In this section, we present several representative examples showing such distinctions. Please note that all of these benchmarks have curated a diverse set of questions. We demonstrate examples only to provide an intuition of the \emph{complexity} of question styles instead of claiming that these benchmarks can be solved with a few techniques.

\begin{figure*}[t]
    \centering
    \includegraphics[width=0.75\textwidth]{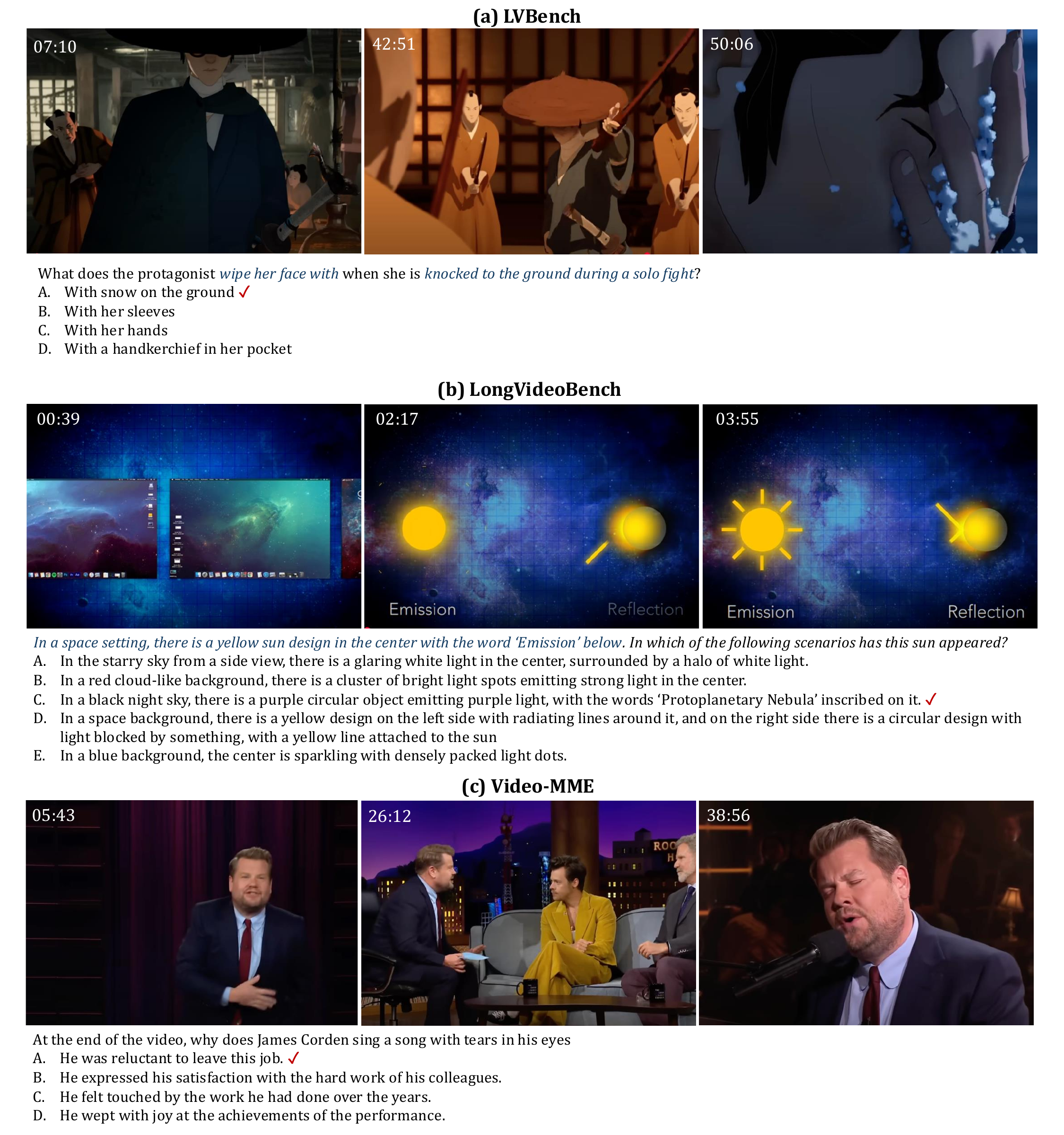}\vspace{-3mm}
    \caption{\textbf{Examples of Different Long Video Benchmarks.}}
    \label{fig:benchmark_examples}
\end{figure*}

As in Fig.~\ref{fig:benchmark_examples}, the LVBench~\cite{wang2024lvbench}, LongVideoBench~\cite{wu2024longvideobench}, and Video-MME~\cite{fu2024video} have different question styles. 

\mypar{LVBench.} For LVBench, the model has to localize the scene of ``solo fight'' correctly and understand the meaning of ``knock down'' and ``wipe face'' to answer the question. Notably, the model has to integrate the contexts of the video and speculate the ``protagonist'' first to execute this task. 

\mypar{LongVideoBench.} Although both require precise localization, LongVideoBench is different from LVBench. LongVideoBench provides explicit and accurate visual cues for the model to localize the object, but the model has to propagate such information across the temporal axis to answer the question. Compared with LVBench, LongVideoBench emphasizes models' visual detail perception and temporal association abilities.

\mypar{Video-MME.} Unlike the above two benchmarks, many questions in Video-MME are not about a specific event. Instead, they are more interpretative, similar to the impression of a human after watching the videos.

We hope the above analysis reveals the spectrum of long video understanding and existing benchmarks.

\section{Prompts and Implementation Details}
\label{sec:prompts}

\subsection{Captioning Prompts}
\label{sec:caption_prompts}

We describe the detailed steps and prompts for our dense captioning of the video (Sec.~\ref{sec:captioning}). All the datasets share the same captioning prompts.

\mypar{Map: Dense Scene Captioning.} As in Sec.~\ref{sec:dense_captioning}, we let each short video segment produce its dense captioning, involving the following three map steps -- all the video segments are independent within each step to support parallel inference: 
\begin{enumerate}
    \item We split each 10s video segment into individual scenes and check if the first scene of a segment can be merged with the last frame of the previous segment. Scene splitting prompts are in Table~\ref{tab:prompts_scene_split}, and the ``Scene Merging'' prompts are in Table~\ref{tab:prompts_scene_merge}.
    \item We identify the salient characters and use them to generate the dense captions in each video segment. The prompts are in Table~\ref{tab:prompts_character_selection}.
    \item With the selected characters, we generate the dense captions of each scene with the prompts in Table~\ref{tab:prompts_dense_description}.
\end{enumerate}

\mypar{Reduce: Consistent Characters and Objects.} As in Sec.~\ref{sec:dense_captioning}, our additional ``Reduce'' step enhances consistency by merging the repeated characters into unified names. It involves the following steps:
\begin{enumerate}
    \item We iteratively check if the characters from two video segments overlap with the prompts in Table~\ref{tab:prompts_character_merge}.
    \item After assigning new names to all the characters/objects, we modify the old names in the original dense captions with the prompts of Table~\ref{tab:prompts_caption_modify}.
\end{enumerate}

\begin{table*}
\scriptsize
\ttfamily
\begin{tabularx}{\textwidth}{X}
\#\# Context

You will be given a few continuous screenshots of the video corresponding to approximately 10 seconds of video duration, and provide detailed, faithful, and accurate analysis of this video segment.
The objective of this analysis is to group the video into short segments based on the contents for the sake of captioning and user question answering.

\\

\#\# Instructions

To perform the analysis of decomposing a video into shorter parts, let's do it step by step.

1. Based on the provided frames of this video segment, please describe the contents of the video segment briefly and accurately. You should cover each action and event in the clip. The description should be detailed, faithful, and accurate. It should come with a header: "[1. Description]:".

2. Based on your description, please answer the following question: "Is this video segment a single scene or a combination of multiple scenes?" The definition of a scene is a single, self-contained, and continuous event that could be easily summarized into one sentence by a human. Your answer should come with a header: "[2. Single:]"

3. If the answer to the previous question is "no", please provide the index of frame(s) separating the scenes from the given frame. Your answer should come with a header: "[3. Frames]:" and in the format of a list of integers.

\\

\#\# Example

Your response should be in the following format:

[1. Description]: This video shows ...

[2. Single: yes/no]: No.

[3. Frames]: [5, 9]

\\

Please pay special attention to:

- The precise localization of the frames is very important for downstream tasks.

- The summarization at the scene level should be consistent with the frames you provided. For instance, the number of scenes should be one more than the number of frames in the list. If you provide 0 frames since the images display a consistent scene, you will give 1 summary; If you provide 1 frame, there should be 2 summaries; if you provide 2 frames, there should be 3 summaries, etc.

Now you will be presented the video frames, please perform the analysis carefully.
\end{tabularx}
\vspace{-3mm}
\caption{``Scene Splitting'' prompts for the Captioning stage (Sec.~\ref{sec:caption_prompts})}
\label{tab:prompts_scene_split}
\end{table*}
\begin{table*}
\scriptsize
\ttfamily
\begin{tabularx}{\textwidth}{X}
You are going to help with determining if a short video segment is a consistent scene. You will be given a few continuous screenshots of the video clip, and provide detailed, faithful, and accurate analysis of this video segment.

\\

Your objective is simple: if *the video clip starting from the second frame* is a consistent scene with *the first frame*. Answer with "yes" or "no".

\\

Now you will be presented the video frames, please perform the analysis carefully.
\end{tabularx}
\vspace{-3mm}
\caption{``Scene Merging'' prompts for the Captioning stage (Sec.~\ref{sec:caption_prompts})}
\label{tab:prompts_scene_merge}
\end{table*}
\begin{table*}
\scriptsize
\ttfamily
\begin{tabularx}{\textwidth}{X}
\# 1. Motivation

You are paritipating in a video captioning task, but you can only watch a few frames of the video and lack a broader context. Therefore, you will use using a character-centric and object-centric visual memory that stores the key characters and objects in the video.
Your objective is to identify the potential key characters and objects from the video, that could be influential, and organize them into a visual memory for downstream tasks.

\\

\# 2. Input and Output

You will be given the following inputs:

\#\# 2.1 Input

You will have sveral sparsely sampled frames of the current video clip.

\#\# 2.2 Output

Your output will have the following format:

[1. Appeared Characters]: You will return a list of the names of the characters and objects that appeared in the current scene, from the visual memory. Strictly follows the format: [NAME1, NAME2, ...]

[2. Character Details]: You will return the details of the characters that appeared in the current scene. Each item should contain the name of the character, a representative frame of the character, and a description about how to identify the character in the frame. Format is:

[Visual Memory 1:] [[NAME: name], [DESCRIPTION: description], [FRAME: index of the selected frame to display this character]] [Visual Memory Ends]

[Visual Memory 2:] [[NAME: name], [DESCRIPTION: description], [FRAME: index of the selected frame to display this character]] [Visual Memory Ends]
...

Guidelines:

1. NAME should be a a general name, such as person\_a, person\_b, person\_c, object\_a, etc. Try to be rigorous and faithful to the video without making assumptions.

2. DESCRIPTION should be a short description of the character's and object's appearance and properties, especially how to uniquely identify the character or object from the representative frame.

3. FRAME should be the index of the frame that best represents the character or object in the scene, favorably the most salient frame showing the front face of the character. It should start from 0.

\\

\#\# 2.3 Example Output:

[1. Appeared Characters]: ["person\_a", "person\_b", "dog\_a"]

[2. Character Details]: 

[Visual Memory 1:] [[NAME: person\_b], [DESCRIPTION: a man with short hair and glasses in the frame], [FRAME: 10]] [Visual Memory Ends]

[Visual Memory 2:] [[NAME: person\_c], [DESCRIPTION: a woman with long hair and a blue dress in the frame], [FRAME: 20]] [Visual Memory Ends]

[Visual Memory 3:] [[NAME: dog\_a], [DESCRIPTION: a dog with standing beside the man with short hair], [FRAME: 10]] [Visual Memory Ends]

\\

\# 3. Guidelines

This is not an easy task, please make sure to use your advanced reasoning ability and check every item and step carefully. The following guidelines are very important for you to finish this task:

1. Please imagine yourself as a human watching the video, trying to perceive the salient things from the video and understanding the deeper plots of the video.

2. When you are selecting the characters for the visual memory, please be picky: 

(a) Only select the characters and objects that you believe are salient and could significantly influence the plot. It could be a person in the movie, an animal in the documentary or cartoon, etc. Make your best judgements.

(b) Only include a character if it is displayed saliently with great emphasis. Be conservative if you cannot identify the character clearly. Better to be safe than sorry.

3. Format is very important. Please keep the strings in identical formattings to ensure smooth post-processing.

4. Please make sure the [1. Appeared Characters] and [2. Character Details] are consistent.

\\

\# 4. Your Job

Now your job begins.
\end{tabularx}
\vspace{-3mm}
\caption{``Character Selection'' prompts for the Captioning stage (Sec.~\ref{sec:caption_prompts})}
\label{tab:prompts_character_selection}
\end{table*}
\begin{table*}
\scriptsize
\ttfamily
\begin{tabularx}{\textwidth}{X}
\# 1. Instructions

You will be given a few continuous screenshots of a video clip, some potential key characters and objects of the video in your memory, and the caption of the previous scene. Your objective is to generate a caption for current displayed scene. Your analysis will be faithful and accurate to the video.

Input:

1. Visual Memory: the names, representative video frames, and the identifiable properties of the characters and objects in the visual memory.

2. Previous Caption: the caption of the previous scene.

3. Video Frames: the few continuous screenshots of the current video clip.

\\

\# 2. Guidelines

When generating the caption, please follow the guidelines below and solve this problem step by step:

1. First, describe the main content of the current scene briefly.

2. Second, use the visual memory to identify if any characters or objects from the visual memory appear in the current scene. If so, please list their name out.

3. Third, describe the scene in detail, including the characters, their actions, the objects, the properties of the characters and objects, the environment, and other types of contents, etc.

Some more detailed tips:

1. When generating the captions, please take the previous scene as contexts and pretend that you are watching a video continuously. The goal is that a human should read your captions and feel like watching a continuous video.

2. When generating the captions, please be faithful to the video and make logical connections between the scenes. 

3. When you encounter characters, please utilize the information and name from the visual memory if what you see matches the visual memory. For instance, if the visual memory contains a character named "person\_a", you should use <person\_a> to refer to the character in your captions.

Important Rules:

1. The quality of this step is very very very important.

2. I want you to be very detailed and faithful to the video. At least, you should go over the following aspects:

2.1 What are the characters, what are their appearances, what are there clothes, what are their actions, what are their emotions?

2.2 What are the objects, what are their properties, what are their relationships with the characters?

2.3 What are the environments, what are the background, what are the weather, what are the time of the day?

2.4 Are there any text on the screen? What are they?

2.5 If there is anything salient or anything weird, please describe it.

\\

\# 3. Format

Your response should be in the following format:

[1. Brief Description]: ... \# captions, a string

[2. Appeared Characters]: ... \# the format of [NAME1, NAME2, ...], a list of character or object names

[3. Detailed Description]: ... \# the detailed description of the scene, a string

\\

\# 4. Your Job

Now your job begins.

\end{tabularx}
\vspace{-3mm}
\caption{``Dense Captioning'' prompts for the Captioning stage (Sec.~\ref{sec:caption_prompts})}
\label{tab:prompts_dense_description}
\end{table*}
\begin{table*}
\scriptsize
\ttfamily
\begin{tabularx}{\textwidth}{X}
\# 1. Instructions

You will be given two sets of frames captured from a video, describing several characters or objects from the video. Your objective is to find if any character or object appears in both sets. If so, please help me locate the character or object and find the better frame representing the characters and objects.

Input:

1. Set 1: the names, representative video frames, and the identifiable properties of the characters and objects.

2. Set 2: the names, representative video frames, and the identifiable properties of the characters and objects.

\\

\# 2. Guidelines and Tips

This is not an easy task, please make sure to use your advanced reasoning ability and check every item and step carefully. The following guidelines are very important for you to finish this task:

1. Please work on this problem via two steps: (a) check if any items from the first set is repeated with the second set; (b) if so, find the better frame representing the character or object.

2. Please rely on both the video frame information and the identifiable properties to carefully understand the characters and objects.

3. When you are selecting the better frame for an object, please consider the following factors: (a) the frame should be the most salient frame showing the front face of the character; (b) the frame should be the most representative frame showing the character or object.

4. Sometimes the characters or objects are captured from different angles or distances, please make your best judgement to check if they are the same character or object.

\\

\# 3. Output Format

Please strictly follow the format below to ensure smooth post-processing:

[Repeated Characters and Objects]: (Character\_name1\_in\_Set\_1, Character\_name1\_in\_Set\_2, Better\_character\_name1), (Character\_name2\_in\_Set\_1, Character\_name2\_in\_Set\_2, Better\_character\_name2) ...

The answer lists all the repeated characters and objects in the two sets of frames, each tuple contains three items describing the repeated character or object:

1. Character\_name\_in\_Set\_1: the name of the character or object in the first set of frames.

2. Character\_name\_in\_Set\_2: the name of the character or object in the second set of frames.

3. Better\_character\_name: the name of the better character or object that represents the repeated character or object, must be consistent with the name in Character\_name\_in\_Set\_1 or Character\_name\_in\_Set\_2.

An example output should be:

[Repeated Characters and Objects]: (person\_a, person\_b, person\_a), (dog\_a, dog\_b, dog\_b)

\\

\# 4. Your Job

Now you will receive two sets of frames and their character descriptions. Please start your responses with the information provided.
\end{tabularx}
\vspace{-3mm}
\caption{``Character Merging'' prompts for the Captioning stage (Sec.~\ref{sec:caption_prompts})}
\label{tab:prompts_character_merge}
\end{table*}
\begin{table*}
\scriptsize
\ttfamily
\begin{tabularx}{\textwidth}{X}
\# 1. Instructions

You will be given a description of a video clip, which potentially contains some characters. After some analysis, I have decided to change the name of the characters or objects, and your job is to help me modify the descriptions to the new names.

Input:

1. Old Description: the old description of the video clip, containing the fields of Brief Description, Appeared Characters, and Detailed Description.

2. Modified List: a list of characters to be modified in the format of OLD\_NAME -> NEW Name.

Output:

Your output should be the modified description of the video clip strictly following the original format and contents, only with names changed.

\\

\# 2. Guidelines

1. Only change the names, do not change the format or any contents.

2. Please remember to update all the Brief Description, Appeared Characters, and Detailed Description.

3. Keep the names consistent.

4. The format of the characters in Brief and Detailed Description is <NAME>, please follow the same format.

\\

\# 3. Your Job

Now your job begins.
\end{tabularx}
\vspace{-3mm}
\caption{``Caption Modification'' prompts for the Captioning stage (Sec.~\ref{sec:caption_prompts})}
\label{tab:prompts_caption_modify}
\end{table*}

\subsection{Analysis I Prompts}\label{sec:supp_analysis_1}

We describe the prompts for question intention analysis (Sec.~\ref{sec:analysis_1}).

\mypar{Map: Segment Intention Analysis.} We let a standard LLM check the scene-level information and understand the user intentions. Each chunk of captions contains 50 scenes. Its prompts are in Table~\ref{tab:prompts_segment_intention}.
\begin{table*}
\scriptsize
\ttfamily
\begin{tabularx}{\textwidth}{X}
\# 1. Motivation

You will conduct the first step of long video understanding: **perceiving short video segments** and **analyze their relevance to the user's question**. By using short-segment analysis, you can avoid the limitation of the model's context length for long videos.

You will have access to the following information for the current video segment:
\\
1. A question.
\\
2. The frames sampled from the video, each corresponding to a scene in the captions.
\\
3. The captions of the video generated by a video captioning model, decomposed into short scenes representing different video actions. Notably, we have marked the potentially key characters or objects using the format of <NAME>. However, it is not entirely reliable (e.g., missing characters or inconsistent tracking across frames), please use it with reasoning.

\\

\# 2. Output Formats

\\
Please strictly following the output format below, which is important for post-processing.

[1. Reasoning]: ... (Your reasoning process. Please be precise, concise, and clear. Mentioning evidence is any.)

[2. Relevant Segments]: [(t\_start, t\_end), ...]... (List the time range of the video segments that are relevant to the question. Please strictly follow the time information from the captions. if you think a continuous period is necessary for the question, merge them into a single segment. Return an empty list if none of the segments are relevant.)

[3. Confidence Level]: ... (Your confidence level.)

[4. Key Characters]: [(character symnonym in question, identifiable properties or NAME in captions), ...]... (The key characters that are mentioned in the question and how to identify them. Keep the list empty if the question is not related to any characters.)

\\

\# 3. Instructions and Guidelines
\\
\#\# Information Reliability

To principle is to **combine the information from the captions, video frames, and the question (including the options, if any)** to analyze the user's intention. The reliability of the information is:

1. The question: raised by the user, the most important and reliable.

2. Video frames: reliable, but only covers a small portion of the video.

3. Captions: less reliable, but covering more details, especially the "NAME" representing character/object names. You should combine the information from the question and video frames when using the captions.

\\
\#\# Analysis Tips
\\

1. Think carefully about how a short video segment could contribute to long video understanding by paying attention to the question and video segment contents. Some examples are:
\\
    - For question on visual details, you should check if the video segment **contains the scene that the user wants**.
    \\
    - For question on information over a period of time, such as the order or the number of actions, you should reason **whether this segment can contribute part of the analysis**.
    \\
    - For question on the reason or implication of the story/actions in the video, you should check if the video segment **contains the key information** that can help you understand the story/actions.
\\
2. Finding the key video segment is critical. If the user mentions a clear criteria, such as specific character of object, try to use it **precisely** and **rigorously** in your analysis.
\\
3. If the question asks for certain characters in the plot/story, you should potentially localize its NAME in the captions, or clearly specify its appearance properties.
\\
4. Pay attention to the information reliability mentioned above.
\\
5. Imagine yourself watching a video using the sampled frames and the captions.
\\
6. When discussing your analysis, please provide the reasoning process and your confidence level between 1 (almost guessing, no clear evidence of being relevant to the question) to 5 (almost certain, clear evidence of being relevant to the question).
\\
7. If the question should be answered with contexts, for "Relevant Segments", you should include one more scene before and after the most possible scene to increase robustness. For example, if the most possible segment is (10, 20), and its previous and next scenes are (5, 10) and (20, 25), then you should make it (5, 25) so that the contents between two scenes won't be missed.

\\

\#\# Your Input
\\
1. The question: a question coming with options.

2. The frames: a list of frames sampled from the video.

3. The captions: a list of captions decomposed into short scenes representing different video actions. Each caption is the format of "(t\_start, t\_end): caption". Time is represented in seconds.

\\

\# 4. Your Job Starts

\end{tabularx}
\vspace{-3mm}
\caption{``Segment Intention Analysis'' prompts for the Question Intention Analysis stage (Sec.~\ref{sec:supp_analysis_1})}
\label{tab:prompts_segment_intention}
\end{table*}

\mypar{Reduce: Global Intention Analysis.} This step utilizes an LLM to process the segment-level analyses from the previous step and unify them into a condensed video-level analysis. The prompts are in Table~\ref{tab:prompts_video_intention}. The most critical part is explicitly instructing the LLM to conduct video-level reasoning and find the most proper scenes.
\begin{table*}
\scriptsize
\ttfamily
\begin{tabularx}{\textwidth}{X}
\# 1. Motivation

You will conduct **user intention analysis** as a step of long video understanding: what is the question asking about. The questions from the users might be vague or not self-contained. You will complete the question by finding the relevant video segments, characters/objects, or how the short video segments contribute to the long video understanding.

You will have access to the following information:

1. A question.

2. Your analysis of short video segments: **is the video segment relevant to the question?**

Your analysis is the most important information in this step. You will go through the analysis of each segment containing the following parts:

1. Reasoning: ... (your explanation)

2. Relevant Segments: [(t\_start, t\_end), ...]... (The periods that are potentially relevant from your analysis. Time is represented in seconds.)

3. Confidence Level: ... (Your confidence level.)

4. Key Characters: [(character symnonym in question, identifiable properties or NAME in captions), ...]... (The key characters that are mentioned in the question and how to identify them. Could be unreliable.)

\\

\# 2. Instructions and Guidelines

\#\# Objectives

Your goal is to merge the information from separate short video segments into a complete understanding at the video level. Your most critical output for the downstream parts are the "relevant segments" and "key characters". Notably, you will carefully use your reasoning skills to handle the following issues:

1. Segment-level analysis might guess some relevant segments or characters for the question. You should select the most relevant segments and characters based on a video-level understanding, and ignore the less relevant ones.

2. Segment-level analysis might contain contradicting information since they come from separate analyses. You should carefully merge the information from different segments, and provide reliable information for the downstream analyses steps.

3. You should clarify how the results from segment-level can contribute to the long video understanding. For example, do we want to "sum", "merge", or "select" the information from individual segments.

\\

\#\# Output Formats

[1. Reasoning]: ... (Your reasoning process. Please be precise, concise, and clear. Mentioning evidence is any.)

[2. Relevant Segments]: [(t\_start, t\_end), ...]... (List the time range of the video segments that are relevant to the question. Please strictly follow the time information from the analysis provided to you. Merge the scenes if you think a continuous period is necessary for the question.)

[3. Key Characters]: [(character symnonym in question, identifiable properties or NAME in captions), ...]... (The key characters that are mentioned in the question and how to identify them. Keep the list empty if the question is not related to any characters.)

[4. Local or Global]: ... (Whether the question requires combining contexts from different segments to answer. If "yes", then this is a global question. If "no", then this is a local question.)

It is very important to follow the format for the relevant segments section. Every segment should be a format of (t\_start, t\_end), especially the brackets should be "()" and matched.

\\

\#\# Principles and Tips

1. Think carefully about how a short video segment could contribute to long video understanding by paying attention to the question and video segment contents. Some examples are:

    - For question on visual details, you should check if the video segment **contains the scene that the user wants**.
    
    - For question on information over a period of time, such as the order or the number of actions, you should reason **whether this segment can contribute part of the analysis**.
    
    - For question on the reason or implication of the story/actions in the video, you should check if the video segment **contains the key information** that can help you understand the story/actions.

2. Finding the key video segment is critical. If the user mentions a clear criteria, such as specific character of object, try to use it **precisely** and **rigorously** in your analysis.

3. If the question asks for certain characters in the plot/story, you should potentially localize its <NAME> in the captions, or clearly specify its appearance properties.

4. Imagine yourself watching a video using the sampled analysis. Figuring out the flow of the plots is critical.

5. If the question is not really about the **whole video**, do not specify more than 10 relevant segments.

6. You should propose **at least 1 relevant segment**. If you don't think any segment is relevant, return a most likely segment and say "I have low confidence on the relevance of the segments".

\\

\# 3. Your Job Starts

\end{tabularx}
\vspace{-3mm}
\caption{``Global Intention Analysis'' prompts for the Question Intention Analysis stage (Sec.~\ref{sec:supp_analysis_1})}
\label{tab:prompts_video_intention}
\end{table*}

\subsection{Analysis II Prompts}\label{sec:supp_analysis_2}

This section provides the details and prompts for \name's goal-aware analysis (Sec.~\ref{sec:analysis_2}).

\mypar{Map: Goal-aware Scene-centric Analysis.} Based on the information required to answer the question, \name first proposes customized queries for each question and as in Table~\ref{tab:prompts_customized_queries} and apply these questions to the VLMs.

\mypar{Reduce: Answer Generation.} The final step is to combine the results of goal-aware scene-centric analysis with the global intention analysis to generate a final response. The prompts are in Table~\ref{tab:prompts_answer}.

\begin{table*}
\scriptsize
\ttfamily
\begin{tabularx}{\textwidth}{X}

\# 1. Motivation

In this step of long video understanding, you are making preparations for calling vision-language models to analyze sampled video frames. Specifically, you will be given the user's question and a video-level analysis from yourself. Based on such information, you will **propose a question to prompt the vision-language models** to analyze the video frames.

You will access the following information:

1. A question.

2. A video-level analysis from yourself. It contains the following information:

1. Reasoning: ... (Your explanation about which parts of the video are relevant to the question.)

2. Relevant Segments: [(t\_start, t\_end), ...]... (The periods that are potentially relevant from your analysis. Time is represented in seconds.)

3. Key Characters: [(character symnonym in question, identifiable properties or NAME in captions), ...]... (The key characters that are mentioned in the question and how to identify them. Keep the list empty if the question is not related to any characters.)

4. Local or Global: ... (Whether the question requires combining contexts from different segments to answer. If "yes", then this is a global question. If "no", then this is a local question.)

\\

\# 2. Instructions and Guidelines

\#\# Objectives

When thinking about the questions to ask, please pay attention to how the next step will sample the video frames for your questions. In practice, we will use two ways:

1. Local: Sample N video frames for each relevant segment, e.g., 32 frames. In this way, the vision-language models can use your question to check the details of each segment.

2. Global: Sample 1 video frame for each segment, sequentially. In this way, the vision-language models can use your question to check the flow of the plots or conduct reasoning over a longer period of time.

Therefore, you should propose two questions:

1. A local question: what kind of detailed information or evidence should the vision-language models find in each segment?

2. A global question: what kind of reasoning should the vision-language models conduct on a longer time span?

\#\# Output Formats

Please strictly follow the output formats below to propose your questions, so that the downstream parts can easily extract the information:

[1. Reasoning]: ... (Your reasoning process. Please be precise, concise, and clear. Explicitly thinking about what kind of information is missing or important for the question.)

[2. Local Question]: ... (Your question for the local analysis.)

[3. Global Question]: ... (Your question for the global analysis.)

\#\# Principles and Tips

1. Think carefully about how a short video segment could contribute to long video understanding by paying attention to the question and video segment contents. Some examples are:
    
    - For question on visual details, you should check if the video segment **contains the scene that the user wants**.
    
    - For question on information over a period of time, such as the order or the number of actions, you should reason **whether this segment can contribute part of the analysis**.
    
    - For question on the reason or implication of the story/actions in the video, you should check if the video segment **contains the key information** that can help you understand the story/actions.

2. Keep your question concise, clear, and within a few sentences. Do not enumerate or explicitly depending on any time information.

3. Remember to use the options from the original questions, expressed with (A), (B), (C), (D), to think about the best way to distinguish the correct one. It is also important to include the original options as the context for the vision-language models.

4. Use your knowledge of prompting large language models or vision-language models to improve your question.

5. Your output questions should only contain a question and options. Do not include any analyses, speculations, or reasoning into the question. For example, the question should directly start as "Describe ... (A) ..., (B) ..., (C) ..., (D) ..., (E) ...", "What is ... (A) ..., (B) ..., (C) ..., (D) ..., (E) ..." or similar formats.

\\

\# 3. Your Job Starts

\end{tabularx}
\vspace{-3mm}
\caption{``Customized Queries for Perception'' prompts for the Goal-aware Analysis stage (Sec.~\ref{sec:supp_analysis_2}).}
\label{tab:prompts_customized_queries}
\end{table*}
\begin{table*}
\scriptsize
\ttfamily
\begin{tabularx}{\textwidth}{X}

\# 1. Motivation

You are at the last step of long video understanding. You will have the user's question and a series of your analysis to finally answer the user's question.

Before conducting actual analysis, it is important to understand the steps of the previoous analysis that will be presented to you:

1. Video-level User Intention Analysis: You first analyze which parts of the video and what kind of characters are relevant to the user's question. You also think about how each video segment could contribute to the long video understanding.

2. Goal Proposal: To call vision-language models to analyze the video segments, you have proposed two questions for the VLMs to use. The first question is called "local question", used for detailed analysis for each segment, and the second question is called "global question", used for joint analysis and reasoning across multiple segments.

3. Goal-aware Analysis: You will receive the results of the vision-language models' perception for each video segment using the local question and across multiple segments using the global question.

By understanding the previous steps, you will have a good understanding of the meaning of the information provided to you, especially which parts are reliable and informative for answering the user's question.

\\

\# 2. Instructions and Guidelines

1. Think carefully about how a short video segment could contribute to long video understanding by paying attention to the question and video segment contents. Some examples are:
    
    - For question on visual details, you should check if the video segment **contains the scene that the user wants**.
    
    - For question on information over a period of time, such as the order or the number of actions, you should reason **whether this segment can contribute part of the analysis**.
    
    - For question on the reason or implication of the story/actions in the video, you should check if the video segment **contains the key information** that can help you understand the story/actions.

2. Carefully consider whether the analysis at local segments or across multiple segments is more important for answering the user's question.

3. With the information provided to you, imagine youself as a human watching the video. Figuring out the flow of the plots is critical.

4. It is possible that some information is vague or contradicting each other. You should utilize advanced reasoning skills to resolve the contradictions. Some very useful principles are:
    
    - If the user has mentioned a specific criteria, try to use it **precisely** and **rigorously** in your analysis.
    
    - Try to utilize the confidence levels provided in the answers.
    
    - Always thinking about your strategy: how the analysis at local segments or across multiple segments could contribute to the long video understanding. For example, do you combine the pieces of information together, summing some numbers, or picking the best segment to answer the question?
    
    - Humans have a limited memory. Always prioritize the most salient information.

5. Pay attention to the time information. They might provide additional correspondence information across different segments and analyses.

\\

\# 3. Output Format

Please provide your answer in the following format:

[1. Reasoning]: ... (Your advanced reasoning based on the information above.)

[2. Answer]: A captal letter from A to E (If you cannot find a correct answer, please make a guess from A to E based on the information you have. To ensure correct post-processing, please strictly use this format. Do not add any characters or spaces.)

\# 4. Your Job Starts

--------------------------------

\end{tabularx}
\vspace{-3mm}
\caption{``Answer Generation'' prompts for the Goal-aware Analysis stage (Sec.~\ref{sec:supp_analysis_2}).}
\label{tab:prompts_answer}
\end{table*}

\subsection{Datasets}
\label{sec:supp_dataset}

\mypar{LVBench Videos.} We clarify the unavailable videos from LVBench, as mentioned in Sec.~\ref{sec:datasets}. LVBench requires users to download from YouTube with provided links to protect the copyright. As of March 1st, 2025, 4 videos are no longer available on YouTube, so we cannot evaluate them. Their IDs are: \href{https://www.youtube.com/watch?v=28CIeC8cZks}{28CIeC8cZks}, \href{https://www.youtube.com/watch?v=idZkam9zqAs}{idZkam9zqAs}, \href{https://www.youtube.com/watch?v=QgWRyDV9Ozs}{QgWRyDV9Ozs}, \href{https://www.youtube.com/watch?v=gXnhqF0TqqI}{gXnhqF0TqqI}. After filtering them out, we still have 1,492 out of 1,543 questions.

\mypar{LVBench Ablation Subset.} We select the first video of each category from LVBench (cartoon, live, self-media, documentary, TV, and sports) and form a subset for the ablation study, as mentioned in Sec.~\ref{sec:datasets}. Th six selected videos are: \href{https://www.youtube.com/watch?v=Cm73ma6Ibcs}{Cm73ma6Ibcs}, \href{https://www.youtube.com/watch?v=TiQBTesZUJQ}{TiQBTesZUJQ}, \href{https://www.youtube.com/watch?v=t-RtDI2RWQs}{t-RtDI2RWQs}, \href{https://www.youtube.com/watch?v=hROKtPqktO8}{hROKtPqktO8}, \href{https://www.youtube.com/watch?v=rSE2YPcv89U}{rSE2YPcv89U}, and \href{https://www.youtube.com/watch?v=CgvJqGxzRfE}{CgvJqGxzRfE}. They consist of 98 questions in total.

\mypar{Breadth Benchmarks.} As mentioned in Sec.~\ref{sec:datasets}, we utilize several long video benchmarks in addition to LVBench~\cite{wang2024lvbench} to provide comprehensive evaluation. However, we evaluate on their subsets due to limited computation resources. (1) LongVideoBench~\cite{wu2024longvideobench}. We evaluate the $(900, 3600]$ duration group of the validation set of LongVideoBench. There are 188 videos and 564 questions in total. In the comparison, the accuracies of the VLMs come from Table 5 of the LongVideoBench paper. (2) EgoSchema~\cite{mangalam2023egoschema}. We evaluate on the validation set of EgoSchema, which contains 500 videos and questions. The performance mainly comes from the Table 2 of VCA~\cite{yang2024vca}. (3) Video-MME~\cite{fu2024video}. Our evaluation follows the long video subset of Video-MME, under the setting of not using subtitles. This set contains 300 videos and 900 questions in total. The performance of models comes from the \href{https://video-mme.github.io/home_page.html#leaderboard}{official leaderboard} as of March 1st 2025.

\subsection{Analytical Experiments Details}

\subsubsection{Baseline Evaluation}
\label{sec:gemini_flash_prompts}

As mentioned in Sec.~\ref{sec:implementation}, we evaluate Gemini-2.0-Flash on the long video benchmarks. For the 30min to hour ones, including LVBench, LongVideoBench, and Video-MME, we follow the standard setting of uniformly sampling 256 frames from each video. For EgoSchema, whose videos are 3min, we uniformly sample 128 frames for evaluation. With LVBench frequently asking about events of specific timestamps, we further provide each frame's seconds as interleaved images and texts. Since LongVideoBench's questions are commonly related to the subtitles, we provide the subtitles of sampled frames as the contexts to the VLM. The prompts used for evaluation are in Table~\ref{tab:prompts_gemini}. 

\begin{table*}
\scriptsize
\ttfamily
\begin{tabularx}{\textwidth}{X}

You are a helpful assistant with the ability of watching videos and answering the questions raised by human users. You will process a few continuous screenshots of the video, and answer the questions raised by human users. If you encounter any issues that you cannot answer the question, please pick the most possible answer from the options.

When you answer, please follow the format of:
[1. Reasoning]: ... (Why you choose this answer)
[2. Answer]: ... (The answer you choose, from A, B, C, D)

Important: If you cannot answer the question, please pick the most possible answer from A, B, C, D, E. Do not leave it blank or select other options.
\end{tabularx}
\vspace{-3mm}
\caption{The prompts used for evaluating Gemini-2.0-Flash (Sec.~\ref{sec:gemini_flash_prompts}).}
\label{tab:prompts_gemini}
\end{table*}

\subsubsection{Ablation Study on Finding Relevant Segments}
\label{sec:supp_find_relevant_seg_details}

This section describes more details about the analytical experiments conducted in Sec.~\ref{sec:ablation_intention_analysis}, where we compare the question intention analysis of \name with a multi-modal retriever, MM-Embed~\cite{lin2024mm}.

\mypar{Types of Questions.} Since LVBench's annotations for ``summarization'' and ``reasoning'' questions might specify long ranges, our evaluation mainly focuses on the question types with precise intervals: key information retrieval, event understanding, entity recognition, and temporal grounding. On our subset for analysis, this results in 64 questions.

\mypar{MM-Embed's Retrieval.} Following the practice of VideoAgent~\cite{wang2024videoagent}, every video frame is encoded by concatenating its image contents with a timestamp since some questions are related to specific seconds. In addition, every question is encoded along with its options as some questions do not contain specific contexts. Since \name might propose multiple candidate scenes, we let the retriever select the same number of top-k candidates for a fair comparison. Finally, every question searches its relevant frames via maximum inner-product between the question and video frame embeddings. 

\end{document}